\documentclass[preprint, 1p]{elsarticle} 
\usepackage{graphics}
\usepackage{graphicx}
\usepackage{epsfig,epsf}
\usepackage{amssymb,mathrsfs,amsmath}
\usepackage{cite}
\usepackage{amssymb,amsfonts}
\usepackage{rotating,booktabs,caption}
\usepackage{xcolor}
\usepackage{url}

\usepackage[normalem]{ulem}
\usepackage{caption}
\usepackage{subcaption}
\usepackage[utf8]{inputenc}

\usepackage{cite}
\usepackage{textcomp}
\begin{document}

  \title{Effect of Balancing Data Using Synthetic Data on
 the Performance of  Machine Learning Classifiers for Intrusion 
 Detection in Computer Networks}  

\author[]{Ayesha S. Dina}
\ead{adi252@uky.edu}
\author[]{A. B. Siddique}
\ead{siddique@cs.uky.edu}
\ead[url]{https://www.cs.uky.edu/~siddique/}
\author[]{D. Manivannan\corref{cor1}}
\ead{mani@cs.uky.edu}
\ead[url]{http://www.cs.uky.edu/~manivann}

\affiliation{organization={University of Kentucky},
addressline={Department of Computer Science},
  city={Lexington, Kentucky},
postcode={40508},
country={USA}}
\cortext[cor1]{Corresponding author}

\begin{abstract}

Attacks on computer networks have increased significantly in recent days,
due in part to the availability of sophisticated tools for launching such
attacks as well as thriving underground cyber-crime economy to
support it. Over the past several years, researchers  in academia and
industry used machine learning (ML) techniques to design and implement
Intrusion Detection Systems (IDSes) for computer networks. Many of these researchers used datasets
collected by various organizations to train ML models for predicting
intrusions. In many of the datasets used in such systems, data are
imbalanced (i.e., not all classes have equal amount of samples). With
unbalanced data, the predictive models developed using ML  algorithms
may produce unsatisfactory classifiers which would affect accuracy in
predicting intrusions. Traditionally, researchers used over-sampling
and under-sampling for balancing data in datasets to overcome this
problem.  In this
work, in addition to over-sampling, we also use a synthetic data
generation method, called  Conditional Generative Adversarial Network
(CTGAN), to balance data and study their  effect on various ML
classifiers. To the best of our knowledge,  no one else has used CTGAN
to generate synthetic samples to balance intrusion detection datasets. Based on extensive
experiments using a widely used dataset NSL-KDD,  we found that
training ML models on  dataset balanced
with  synthetic samples generated by CTGAN
increased prediction accuracy by  up to $8\%$,  compared to training
the same ML models over unbalanced data. Our experiments also show
that the accuracy of   some ML models trained over data balanced with
random over-sampling decline compared to the same ML models trained over
unbalanced data.

\end{abstract}


\maketitle

\section{Introduction}
 
There has been a significant increase in the number of intrusions into computer networks over the past few years due in part to sophisticated tools to launch such attacks as well as a thriving underground economy to support such attacks~\citep{DINA2021100462}.
According to a 2017 report~\citep{lastlinecostanalysis}, data breaches cost an average of \$141 per record.
It is estimated that 60\% of small businesses that suffer a data breach will cease operations within six months. 
Symantec's Internet Security Threat Report for 2017 indicated that
the number and intensity of attacks were significantly higher than those in previous years~\citep{khraisat2019survey} -- zero-day attacks totaled more than three billion in 2016.
Traditional tools such as firewalls
can not cope with these sophisticated attacks. In order to fight
network intrusions, hardware and software tools can be
installed to continuously monitor the network. 

James Anderson
published a report on the need for detecting network intrusions in
computer systems~\citep{Anderson-1972} in
1972~\citep{Bridges-2020}. Since then, several intrusion
detection systems (IDSes) have been proposed and implemented.
These systems  can be further  classified as
host-based, network-based, and hybrid~\citep{Stallings-2018}. System
architectures can be centralized, distributed, or hybrid, based on how
intrusion/attack events are collected, processed, and  acted
upon. Certain approaches are superior to others based on factors such
as cost, performance, and other metrics. 
These systems can be further classified based on the
 techniques used for intrusion detection --  signature-based or
 anomaly-based. A Signature-based IDS detects  attacks based on
 the signatures of previously known attacks. These IDSs cannot detect
 zero-day attacks. 
 In contrast, anomaly-based IDSes are capable of  detecting zero-day attacks by modeling users' behaviors.
In the training phase of an anomaly-based
 approach, legitimate users' behaviors are first collected and analyzed in order
 to build a model of legitimate users' behavior. The model is
 then used to  determine whether
 the current observed behavior is that of legitimate user  or not.  Some methods used for such
 classification  are~\citep{Stallings-2018}: {\bf Statistical
   approach}: classification is based on univariate, multivariate, or
 time-series models. {\bf Knowledge based approach}: expert system is
 used to model legitimate behavior according to a set of   rules. {\bf
   Machine learning based approach}: automatically classified  based
 on some clustering algorithms. However, anomaly-based IDSes often
 generate more false positives and signature-based IDSes generally
 generate more false negatives.

 \begin{figure}[htbp]
\centerline{\includegraphics[width=0.9\textwidth]{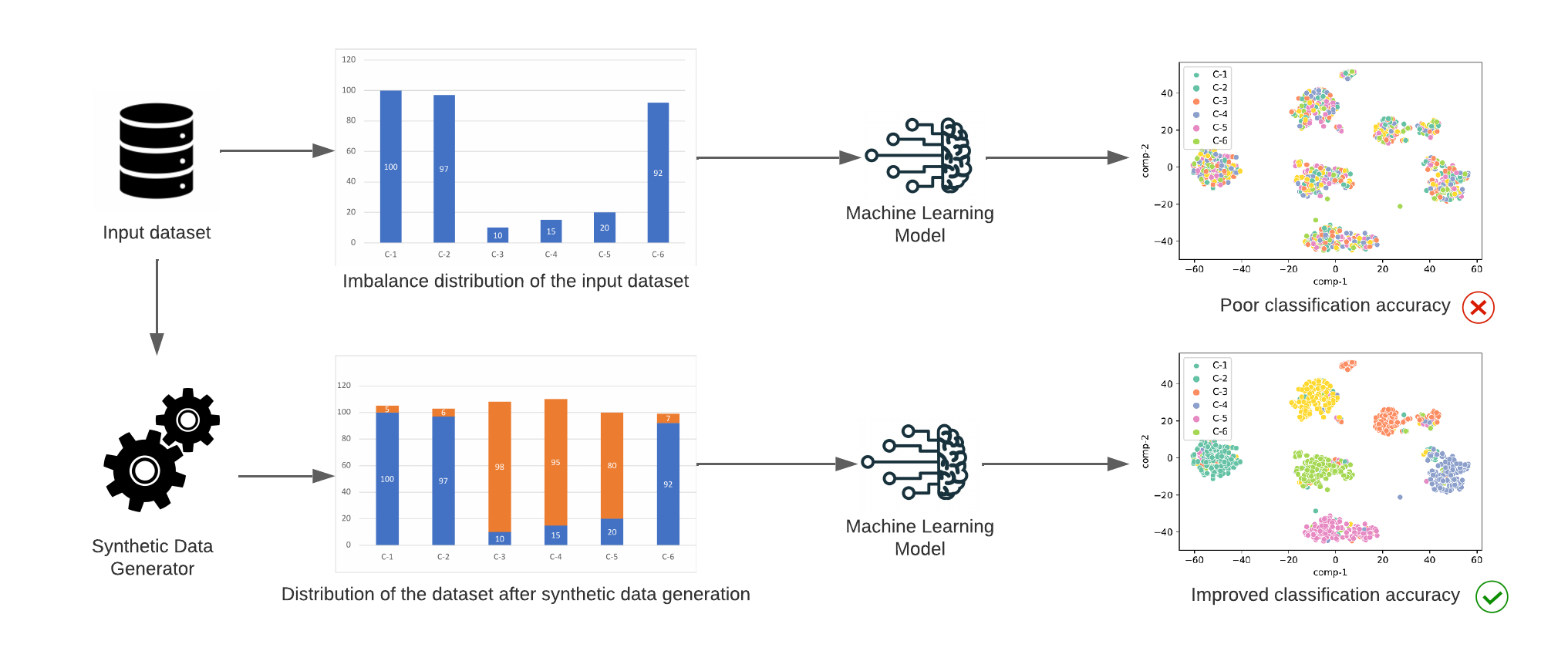}}
\vspace{-15pt}
\caption{
ML models show poor performance on imbalanced datasets -- no matter what type of machine learning model is employed. Synthetic data generation for minority classes can overcome the data imbalance issue, thus may yield improved classification performance of ML models.
}
\label{fig: intro}
\vspace{-8pt}
\end{figure}

ML based IDSes have been extensively studied in the literature. 
For example, following ML approaches have been
tested by various researchers for intrusion detection: Artificial
Neural Networks, Association Rules and Fuzzy Association Rules,
Bayesian Networks, Clustering, Decision Trees, Evolutionary
Computation, Hidden Markov Models, Inductive Learning, Na\"ive Bayes,
Sequential Pattern Mining, and Support Vector
Machine~\citep{Buczak-2016, saranya2020performance, buczak2015survey}. 
In many
of the  datasets used in such studies, data are not balanced.  
That is, 
instances of one class surpass those of another
class~\citep{abd2013review}. The classes that have a high frequency of
instances are called majority classes, while the classes that have a
low frequency of instances are called minority classes. The ratio of samples
between a minority class and 
those of
in  a majority class  may be
as small as  1:100, or as large as 
1:1000, or even larger~\citep{chawla2004special}.  Figure~\ref{fig: intro} highlights that imbalance in the dataset may result in poor
 performance of ML classifiers. Many of
the researchers have (i)  ignored this problem, (ii) balanced  data
using over-sampling (e.g., by randomly replicating
minority class samples) or under-sampling (e.g., by randomly eliminating
majority class samples) techniques.  Over-sampling and under-sampling
help in balancing data. However,  since  the new
samples  added under over-sampling are exact copies of the original
samples, it may cause overfitting. Similarly, since
random data are eliminated from majority classes in under-sampling, the dataset may
become too simple, with the same number of features and too little data to build
an effective model, resulting in underfitting problem.
In general, an overfit model has low bias and high variance, while an underfit model has high bias and low variance.

 In this paper, to study the effect of balancing data on the
 performance of ML classifiers, we use
 (i) the most commonly used  random over-sampling, and  (ii) synthetic
 data generated using the  Conditional Generative Adversarial Network
 (CTGAN)~\citep{xu2019modeling} for balancing data. 
CTGAN exploits a conditional generative adversarial network, learns
from input data (i.e., both discrete and continuous features), and
generates high-fidelity synthetic samples. We generate synthetic
samples for all classes after training CTGAN on the training
dataset. By discarding samples of the majority classes and keeping
samples of the minority classes, we eliminate the data imbalance
issue. It is important to emphasize that the new synthetic samples
generated by CTGAN are not copies of the instances in the original
dataset but look-alike instances. 
To the best of our
 knowledge, this is the first time CTGAN has been used to generate
 synthetic data for balancing data related to intrusion
 detection. 
 
We studied the effect of  synthetic data generation
approaches on
 the performance of various state-of-the-art ML classifiers through extensive
 experiments. We also compare how various  ML classifiers  perform
 on the original dataset.  
  We use NSL-KDD (Network Socket Layer-Knowledge Discovery in Database) dataset in our experiments --
 a widely used dataset for
 studying intrusion detection. Specifically,  we use the following
 state-of-the-art ML classifiers for conducting the experiments:
 Decision Tree (DT), Support Vector Machine (SVM), Random Forest (RF),
 Naive Bayes (NB), Feed Forward Network (FNN), Long Short Term Memory
 (LSTM), and Convolutional Neural Network (CNN).  Our focus is on
 multi-class classification rather than binary classification -- a challenging problem setup. A
 multiclass classification makes it possible to evaluate the
 performances of various models relative to different types of intrusions.
 Our experimental results show that  on the NSL-KDD dataset, with data
 balanced using the synthetic data generated by CTGAN, some of the
 ML classifiers show an increase in prediction accuracy by as much
 as  $8\%$.
 {\bf Following is a  summary of our contribution in this paper:}
 \begin{itemize}
     \item We show that using improved algorithms for
       generating synthetic data for balancing data
       could improve the  performance of  
       ML classifiers in predicting intrusions in computer networks
       more accurately.
       
     \item We use  CTGAN to generate synthetic samples and balance 
       the training samples in NSL-KDD dataset. To the best of our
       knowledge, this is the first time CTGAN has been used to
       generate synthetic data for balancing data associated with
       intrusion detection. It is noteworthy to mention that CTGAN has
       been used for image augmentation in the literature.

     \item We
       use  state-of-the-art classification techniques such as
       DT, SVM, RF, NB, FNN, LSTM, and CNN to classify the balanced
       input dataset. A wide range of evaluations 
       show that ML classifiers trained on NSL-KDD dataset, with
       data balanced using
       CTGAN generated samples,  perform better than the same
       ML classifiers trained  on original NSL-KDD dataset, with data balanced
       using random over-sampling.
 \end{itemize}
 
 The rest of the paper is organized as follows. In
 Section~\ref{methods}, we present our proposed approach 
 as well as discuss the details of the  ML classifiers used for
 evaluation.  In Section~\ref{results}, we present our experimental setup and results. In  Section~\ref{related-work}, we
 discuss related works, and
 Section~\ref{conclusion} concludes the  paper.

\section{Proposed Approach and Classification Methods Used}\label{methods}

In this section, we discuss how we model data, preprocess data, and use
CTGAN to generate synthetic data to  balance data in the NSL-KDD dataset.  Then, we discuss various
ML classifiers that we used in our experimental evaluation.

\subsection{Modeling Data}
We model the input data as  a two-dimensional matrix $X = (x_1, x_2,
x_3...., x_N)$,  where $x_i \in \mathbb{R}^D$ $(1\leq i \leq N)$ is a
vector with $D$ dimensional network feature space. Each $x_i$ $(1 \leq
i \leq N)$ is associated with a label $y_i$  and $y_i \in \{1,
... L\}$. In our  case, $N$ is the number of samples in the dataset
and $L$ is the number of distinct attack categories. The feature
vectors are mapped to labels by a function $Y=f(x)$ that is
unknown. As part of supervised learning, the training dataset is used
to obtain an estimate of $f$. This estimated function is referred to
as $\hat{f}(x)$. The goal is to make $\hat{f}(x)$ as close as possible
to $f(x)$. 
\subsection{Preprocessing of Data}

We transform all the categorical variables into numerical variables
during the preprocessing step. For this transformation, we used label
encoding~\citep{hasan2019attack,mottini2016relative}. During this
process, each label of a categorical feature is assigned a unique
numerical value in alphabetical order. Imagine a two-dimensional
matrix $X$ containing column $C_i$. 
Column $C_i$ contains four categorical labels such as $tcp$, $smtp$, $ftp$ and $http$. These are different types of protocols.
Label encoding will assign a value of $4$, $3$, $1$, and $2$ to the $tcp$, $smtp$, $ftp$ and $http$
labels in alphabetical order.

In the next step, we normalized the input data. In this study, we used
$L_2$ normalization or Euclidean
normalization~\citep{wang2012comparative}. We will use the same input
matrix $X$  and $i^{th}$ feature $C_i$. The feature $C_i$  is
normalized according to Equation \ref{eq:l2-norm}.

\begin{equation}\label{eq:l2-norm}
    C_i = \frac{C_i} {||C_i||_2 } 
\end{equation}
where $$ ||C_i||_2 = {\sqrt{\sum_{k=1}^K c_{k_i}^2}}$$

\noindent Where $C_i = [c_{1_i}, c_{2_i}, c_{3_i},
  .... c_{K_i}]$, a vector of length  $K$, and $||C_i||_2$
is the $L_2$ norm of vector $C_i$.

\subsection{Synthetic Data Generation using CTGAN to Balance Data}
 Data imbalance  occurs when the number of instances in some classes
 is significantly higher than those in other
 classes~\citep{chawla2004special}. Consequently, ML models will be
 overwhelmed by the majority classes (which have higher instances when
 compared to other classes) and ignore the minority classes (which
 have fewer instances). There are several methods to overcome the
 imbalance problem, such as oversampling, undersampling, stratified
 sampling (SS), and so on~\citep{DINA2021100462}, as we mentioned earlier.

In addition to oversampling, we used CTGAN~\citep{xu2019modeling} for generating synthetic data
for balancing data. To generate synthetic
tabular data from original tabular data, CTGAN uses a GAN-based
(generative adversarial network) model.  

CTGAN introduces {\bf{mode specific normalization}}, which allows it
to deal with columns with complex distributions. This procedure
consists of three steps. 
\begin{itemize}
    \item Each continuous column $C_i$ is identified by using a
      variational Gaussian mixture model
      (VGM)~\citep{svensen2007pattern} to determine the number mode
      $m_i$ and fit it in a Gaussian mixture. 
    \item In order to compute the probability density for each mode,
      it computes the value of $c_{i_j}$ in column $C_i$ for $jth$ row.
    \item 
       Then, sample one mode using the  calculated probability
      density and use the sampled mode to normalize the
      value.
\end{itemize}
A new row should be resampled in such a way that all categories from
the columns are equally distributed at the time of training so that it
can be used to capture the actual distribution of data during testing.
Let $k$ be the value of the $i^{th}$ column $C_i$.
Let $\hat{r}$ be a generated sample, and the original value has to be
matched with the generated samples $\hat{r}$ in a way that the
generator can be explained as the conditional distribution of rows
given that particular value at that particular column, where

\begin{equation}
    \hat{r} \sim P_{g}(row|C_i = k)
\end{equation}

One of the most important tasks for the conditional generator is to
learn the real distribution of data, i.e., $P_g(row|C_i = k) =
P(row|C_i = k)$. The following equation can be used to reconstruct the
original distribution. 
\begin{equation}
    P(row) = \sum_{k\epsilon C_i} P_g(row|C_i = k)P(C_i=k)
\end{equation}
In order to achieve this, three methods were introduced: conditional
vectors, generator losses, and sampling-based training. 
Two fully connected hidden layers were used in both the generator and
discriminator of the network architecture in order to capture all
possible correlations between columns. In the generator, batch
normalization and relu activation function are used.

\subsection{Different ML models}
In this subsection, we  discuss different ML classification algorithms
that we used to classify original input data, data balanced with
over-sampling and data balanced with synthetic data generated with CTGAN.
\subsubsection{Decision Tree (DT)}
In many applications, DT has been used to classify different types of
data such as power quality disturbance, parkinson's disease, product
review classification,
etc.~\citep{aich2018nonlinear,syamala2020filter,reges2021decision,kim2021decision,zhao2019power}. 
A DT is tree structure, in which each leaf node represents
a class label and each internal node  is a decision node or a chance
node~\citep{DINA2021100462}. DT constructs a tree by segmenting the
feature space into several subregions. Hence, tree is constructed by
recursively binary splitting of the feature space~\citep{giniequa}.  
Two splitting methods are usually used to split the tree (e.g., cross
entropy, gini index). We used gini index-based
splitting~\citep{peng2018intrusion}. Gini index can be calculated using
Equation~\ref{eq:DT}. 

\begin{equation}\label{eq:DT}
    gini = \sum_{l=1}^L p_l(1-p_l) = 1- \sum_{l=1}^L p_l^2
\end{equation}
\noindent Where $L$ in the number of classes, $p_l$ is a set of items
with class $l \in\{1, 2, 3 .... L\}$

\subsubsection{ Support Vector Machine (SVM)}

SVM model is a renowned machine learning model that can be used for
both classification and regression tasks. It is, however, primarily
used for classification
tasks~\citep{cervantes2020comprehensive,vijayarajeswari2019classification,toledo2019support,chandra2021survey}.
SVM uses  Statistical learning theory  to find the optimal
hyperplane as a decision function in high dimensional
space~\citep{pal2010feature}. We assume a supervised classification
problem, and consider a input set with $N$ vectors from the
d-dimensional feature space $X$. For each vector $x_i$, there is a
target $y_i$~\citep{bazi2006toward}. The goal of SVM is to identify an
optimal hyperplane that maximizes the separation margin. The data are
first mapped to a high dimensional feature space using a kernel
method, i.e., $\phi(X)$. The optimal hyperplane can be defined as 
\begin{equation}
    f(x_i) = w. \phi(x_i) + b
\end{equation}
\noindent Here $f(x)$ represents the discriminant function, $w$ is
weight vector and $b$ is the bias. $b$ minimizes a cost function. The
cost function can be expressed as

\begin{equation}
    \psi(w,\xi) = \frac{1}{2} ||w||^2 + C\sum_{i=1}^N\xi_i 
\end{equation}

\noindent The cost function is obtained by $\xi_i$, which is slack
variable, used for nonseparable data. The constant $C$ is a
regularization parameter to control the shape of the discriminant
function. 
\subsubsection{Na\"{\i}ve Bayes (NB)} 
NB classifiers are a family of probabilistic classifiers
based on Bayes' Theorem. NB classifiers, combined with kernel
density estimation,  can achieve high accuracy levels.
NB is widely used by researchers to solve various classification
problems that arise in their
research~\citep{hartmann2019comparing,deng2019feature,churcher2021experimental}. 
NB classifier is based on conditional
probability~\citep{mukherjee2012intrusion}. The probability of one attribute
does not affect another attribute, given the class label. Therefore, the presence of a
attribute in a class is unrelated to any other attribute. The Naive
Bayes can be written as 
\begin{equation}
    P(L|C) = \frac{P(C|L)P(L)}{P(C)}
\end{equation}
\noindent Where $L$ is the class variable and $C$ is the feature set
$C_1, C_2, C_3..... C_Q$. $ P(L|C)$, $P(C|L)$, $P(L)$, and $P(C)$ are
respectively the posterior probability, probability of feature set
given class, prior probability of class, and prior probability of
feature set.

\subsubsection{Random Forest (RF)}
Due to its simplicity and diversity, RF is also one of the most commonly used algorithms. Regression and classification can both be performed using RF~\citep{komkov2021rf,peng2018design,sheykhmousa2020support,ezuma2019detection}.
RF combines multiple decision trees to make more accurate, stable
predictions. It builds a decision forest based on several decision
trees usually trained with the bagging method. A bagging method, 
based on the concept that combining different learning models,
increases overall performance. In our approach, we used the Gini Index
(Equation~\ref{eq:DT}) to determine how a node in a decision tree
should be split. 

\subsubsection{Feed-forward Neural Network (FNN)}
FNNs have been successfully used for pattern classification,
clustering, regression, association, optimization, control, and
forecasting~\citep{catic2018application,arulmurugan2018early,yang2019feed,statistics-FNN}. FNN
contains one input layer, one output layer, and $H$ number of hidden
layers. Let $W_{h} \in \mathbb{R}^{Q \times P}$, and $W_{o} \in
\mathbb{R} ^ {N \times M}$ be the weight matrices for hidden layer
and output layer respectfully where $Q$ is number of input neurons,
$P$ is the number neurons in a hidden layer and $M$ is the number of
output neurons. Each row of these matrices represents a weight vector
for a neuron. Now we can write the equation of output matrix of a
hidden layer as: 
\begin{equation}
    H = f(XW_{h} + b_h)
\end{equation}
\noindent Where $X= \{x_1,
  x_2, x_3, .... x_N\}$ is the input matrix with $N$  rows, $b_h$ is the bias matrix and $f(.)$ is the
activation function of the hidden layer. 

We can express the equation of the output layer as:
\begin{equation}
    \hat{Y} = g(HW_{o} + b_o)
\end{equation}
\noindent Where $g(.)$ in the activation function of the output layer
and $b_o$ is the bias matrix of the output layer. 
\subsubsection{Long Short Term Memory (LSTM)}
Although LSTM is a recurrent neural network, it is better in terms of
memory than traditional recurrent networks. By memorizing certain
patterns, LSTM is able to perform relatively
better~\citep{nagabushanam2020eeg,jang2020bi,yildirim2019new,saadatnejad2019lstm}.  LSTM can have
multiple hidden layers and as data passes through each layer, the
relevant information is retained and the irrelevant information is
discarded. An LSTM consists of an input gate $i_t$, an output gate
$o_t$, and a forget gate $f_t$. The equations for the LSTM gates at
$t$ time step cam be expressed as: 
\begin{equation}
    i_t = g(W_i[h_{t-1},x_t] + b_i)
\end{equation}
\begin{equation}
    f_t = g(W_f[h_{t-1},x_t] + b_f)
\end{equation}
\begin{equation}
    o_t = g(W_o[h_{t-1},x_t] + b_o)
\end{equation}

\noindent Where $g(.)$ is a activation function of a gate, $W_x$ is
the weight of the corresponding gate, $h_{t-1}$ is the output of the
previous LSTM block, $x$ is the input vector at time $t$, and $b_x$ is
the bias for the respective gate. 
\subsubsection{Convolutional Neural Network (CNN)}
In addition to computer vision, CNNs have shown outstanding performance in many other fields~\citep{lu2021review,dai2020hs,phan2018joint,yu2020simplified}. 
Convolutions
are used in this neural network to transform the input features into
meaningful information, which is then used to build the subsequent
layers of neural network computations. The convolutional layer is used
to extract features, to perform linear operations, and is usually a
combined convolution. In convolution, multiple kernels or filters are
used. A convolutional operation is usually defined as: 

\begin{equation}
    \hat Y = x \times k + b
\end{equation}
\noindent The kernel $k$ has a dimension of $n \times m$. The input
and bias are represented by $x$ and $b$, respectively. The input and
bias have the same dimensions $k$.

\section{Experimental Results}\label{results}
In this section, we  discuss evaluation criteria, dataset used,
experimental setup,
implementation details of classifiers, and
performance of different classifiers.  
\subsection{Evaluation Criteria}
To evaluate the effect  on the performance of various ML-classifiers,
we used the following quantitative metrics: (i)~Accuracy (Acc), (ii) Precision (Pre), (iii) Recall (Rec), and (iv) ${F_1}$
score~\citep{khourdifi2019heart,nguyen2008survey}. {\bf Acc} is
the measure of how well the algorithm correctly predicts the
occurrence of an event.
That is, an event is predicted as normal or type of intrusion.
{\bf Pre} of the prediction refers to how frequently the algorithm actually predicts types of intrusions. {\bf Rec}  refers to the proportion of actual intrusions that the algorithm predicted as intrusions. {\bf $F_{1}$-Score} is equal to the reciprocal of the arithmetic mean of {\bf Pre} and {\bf Rec}, which is the harmonic mean of both variables.

\begin{table}[htbp]
\caption{Performance metrics that are used for comparison}
\begin{center}
\begin{tabular}{c|c}
\hline
\textbf{ Criteria}&{\textbf{Equations}} \\ [0.4ex] 
\hline
Accuracy (Acc) & $\frac{TP+TN}{TP+TN+FP+FN}$  \\ [0.8ex] 

Precision (Pre) & $\frac{TP}{TP+FP}$  \\ [0.8ex] 

Recall (Rec) & $\frac{TP}{TP+FN}$  \\ [0.8ex] 

$F_{1}$ Score &  $2 * \frac{{\bf Pre* Rec}}{{\bf
       Pre + Rec}}$ \\ [0.8ex] 
\hline

\end{tabular}
\label{performance-metrics}
\end{center}
\end{table}

Table~\ref{performance-metrics} summarizes
how  the metrics discussed above are calculated.
We evaluated
various ML models by counting True positives~($TP$), True negatives~($TN$), False positives~($FP$), and False negatives~($FN$).
In this work, all of our classifiers are \emph{multi-class}.

\subsection{Dataset Used for Evaluation}
In this work, we used NSL-KDD (Network Socket
Layer-Knowledge Discovery in Database)~\citep{NSL-KDDdataset} dataset,
which is a widely used dataset in the intrusion detection literature. NSL-KDD dataset contains $41$ features
and it does not contain any 
duplicate records~\citep{tavallaee2009detailed}. 
The dataset has one normal class and four attack type classes.

The four attack types~\citep{ingre2015performance} are:

\begin{itemize}
    \item {\bf Denial of Service (DoS) Attack} – blocking any resources or
      services in a system or network through malicious means.
      \item  {\bf User to Root Attack (U2R)} – the attacker uses a
        normal user account to gain access to the system and exploits
        vulnerabilities to take over the system.
        \item {\bf Remote to Local (R2L) Attack} – unauthorized access
          to a remote 
          system by sending data packets over a network to gain users'
          or root's access to do unauthorized acts. 
    \item  {\bf Probing Attack} - these attacks gather information
      about potential 
      vulnerabilities of target systems so that attacks may be
      launched on them later.

\end{itemize}

It is important to highlight that
there is  significant difference in the sizes of
the training instances of {\bf U2R} and {\bf R2L} classes. The dataset
consists of one set of training data (KDDTrain+) and two sets of
testing data: (i) KDDTest+ and (ii)
KDDDTest21-. 
Table~\ref{NSL-DataDistribution} shows distribution of
instances for various attack  types in the training and testing datasets.

\begin{figure}[t!]
    \centering
    \begin{subfigure}[b]{0.45\textwidth}
        \centering
        \includegraphics[width=\linewidth]{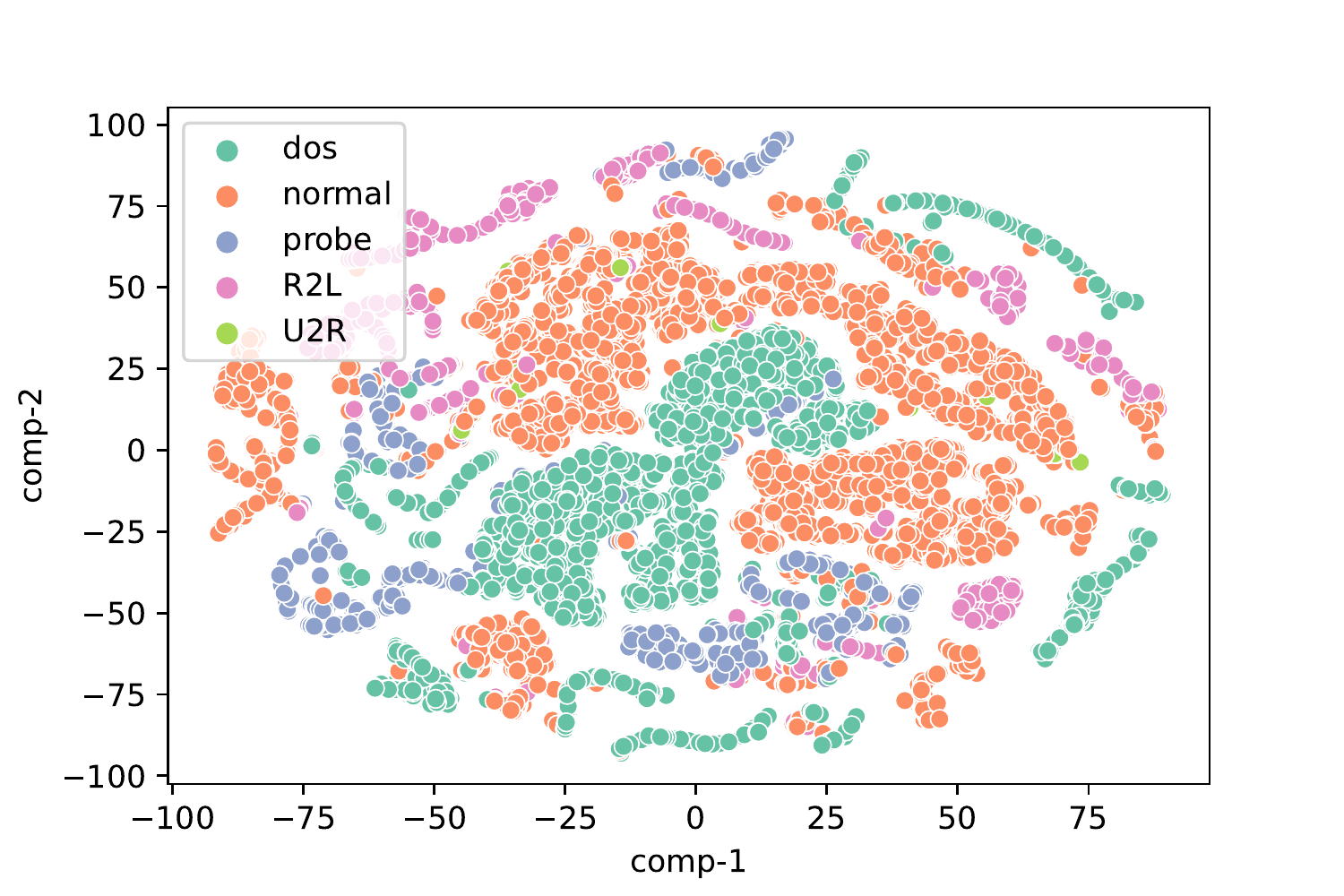}
        \caption{KDDTest+}
    \end{subfigure}
    \hfill
    \begin{subfigure}[b]{0.45\textwidth}
        \centering
        \includegraphics[width=\linewidth]{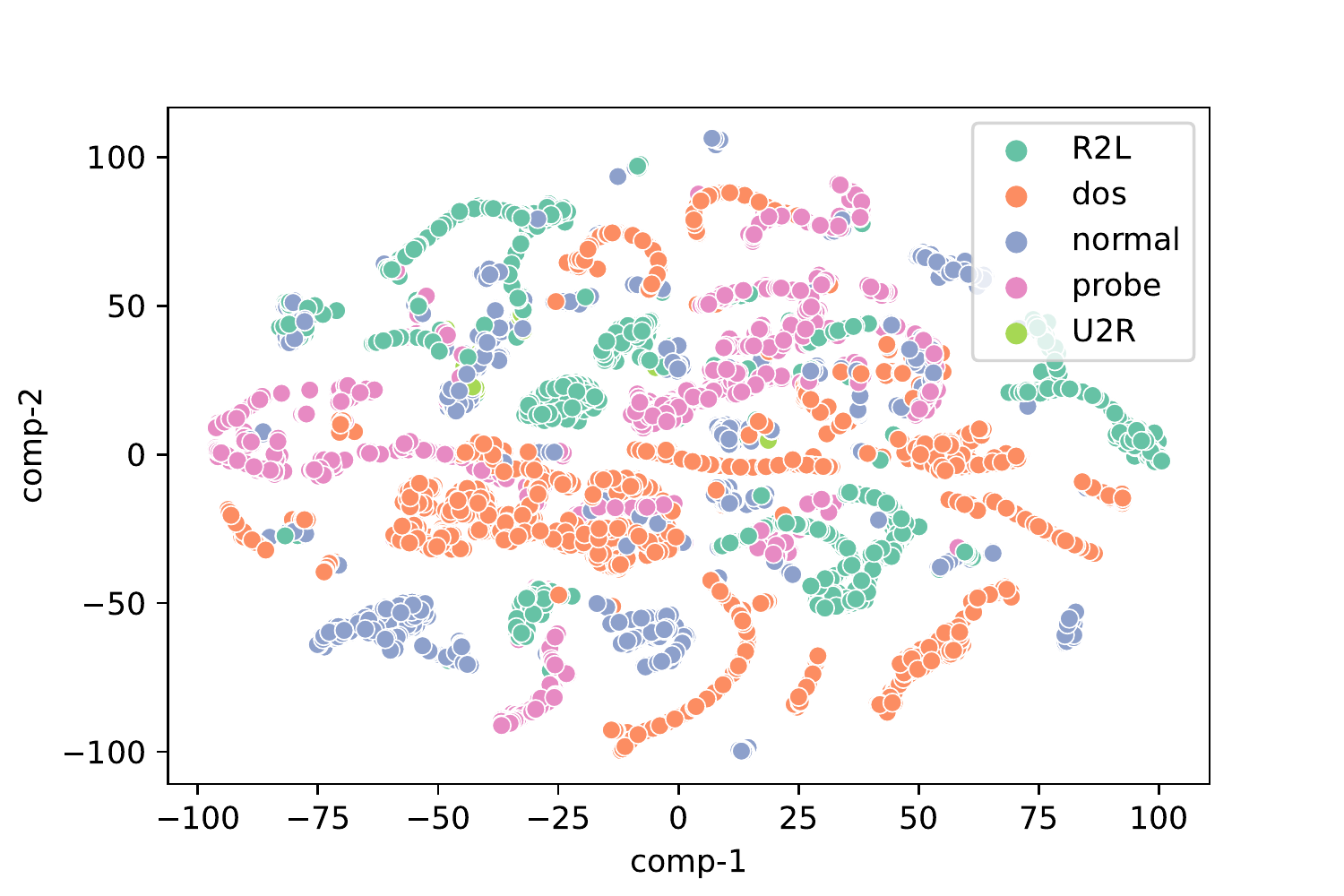}
        \caption{KDDTest21-}
    \end{subfigure} 
    \caption{T-SNE projection of NSL-KDD dataset.
    }
\label{fig: NSL-T-Sne}
\end{figure}

Figure~\ref{fig: NSL-T-Sne} shows partial T-Distributed Stochastic Neighboring Entities (t-SNE) \citep{van2008visualizing} projection of the NSL-KDD datasets. 
Figure~\ref{fig: NSL-T-Sne}(a) presents projections for KDDTest+ test set, whereas KDDTest21- is presented in Figure~\ref{fig: NSL-T-Sne}(b).
We can see in both projections, large portion of the dataset is occupied by normal and DoS class types.

\begin{table}[t!]
\caption{Data distribution in NSL-KDD dataset}
\begin{center}
\begin{tabular}{c|c|c|c|c|c|c}
\hline
{\bf Class}  & {\bf KDDTrain${+}$} & {\bf(\%)} & {\bf KDDTest$+$} & {\bf(\%)} & {\bf KDDTest${21-}$} & {\bf(\%)}\\ [0.5ex] 
\hline
 Normal & $67343$ & $53.5$ & $9711$ & $43.1$ & $13449$ & $53.3$ \\

 DoS  & $45927$ & $36.4$ & $7458$ & $33.1$ &  $9234$ & $36.7$\\

 Probe & $11656$ & $9.3$ & $2421$ & $10.7$  & $2289$ & $9.1$\\

 U2R & $52$ & $0.041$ & $67$ & $0.3$ & $11$ & $0.04$ \\

 R2L  & $995$ & $0.78$ & $2887$ & $12.8$ & $209$ & $0.83$\\
 \hline
\end{tabular}
\label{NSL-DataDistribution}
\end{center}
\end{table}

\subsection{Experimental Setup}
To evaluate different ML models, we conduct  three experiments. 
In the first experiment, we use the original training dataset NSL-KDD
to train ML models. 
In the second experiment, we use random oversampling~\citep{pang2019signature} to balance data and trained
ML models on the balanced dataset.  
In the third experiment, we use CTGAN to produce synthetic samples to balance
data and trained
ML models on the balanced dataset. 
We call these three experiments as  ORG, RndOSamp, and CTGANSamp, respectively.

\subsubsection{Experimental Setup for Original Data (ORG)}
In ORG, we utilize the original training samples from NSL-KDD dataset. The
bar graph  in Figure~\ref{fig: Data distribution}(a) shows distribution of samples under various classes in the original
training samples. We feed these training samples to the
state-of-the-art classifiers without adding any synthetic data to
balance the data.  As shown in Table \ref{NSL-DataDistribution}, the normal and DoS class types are $53\%$ and $36\%$ of training samples, respectively. Whereas, Probe, U2R, and R2L are approximately $9\%$, $0.04\%$, and $0.83\%$ of training samples, respectively.  

\subsubsection{Experimental Setup for Original Data Balanced  with Random Oversampling (RndOSamp)}

\begin{figure}[t]
    \centering
    \begin{subfigure}[b]{0.30\textwidth}
        \centering
        \includegraphics[width=\linewidth]{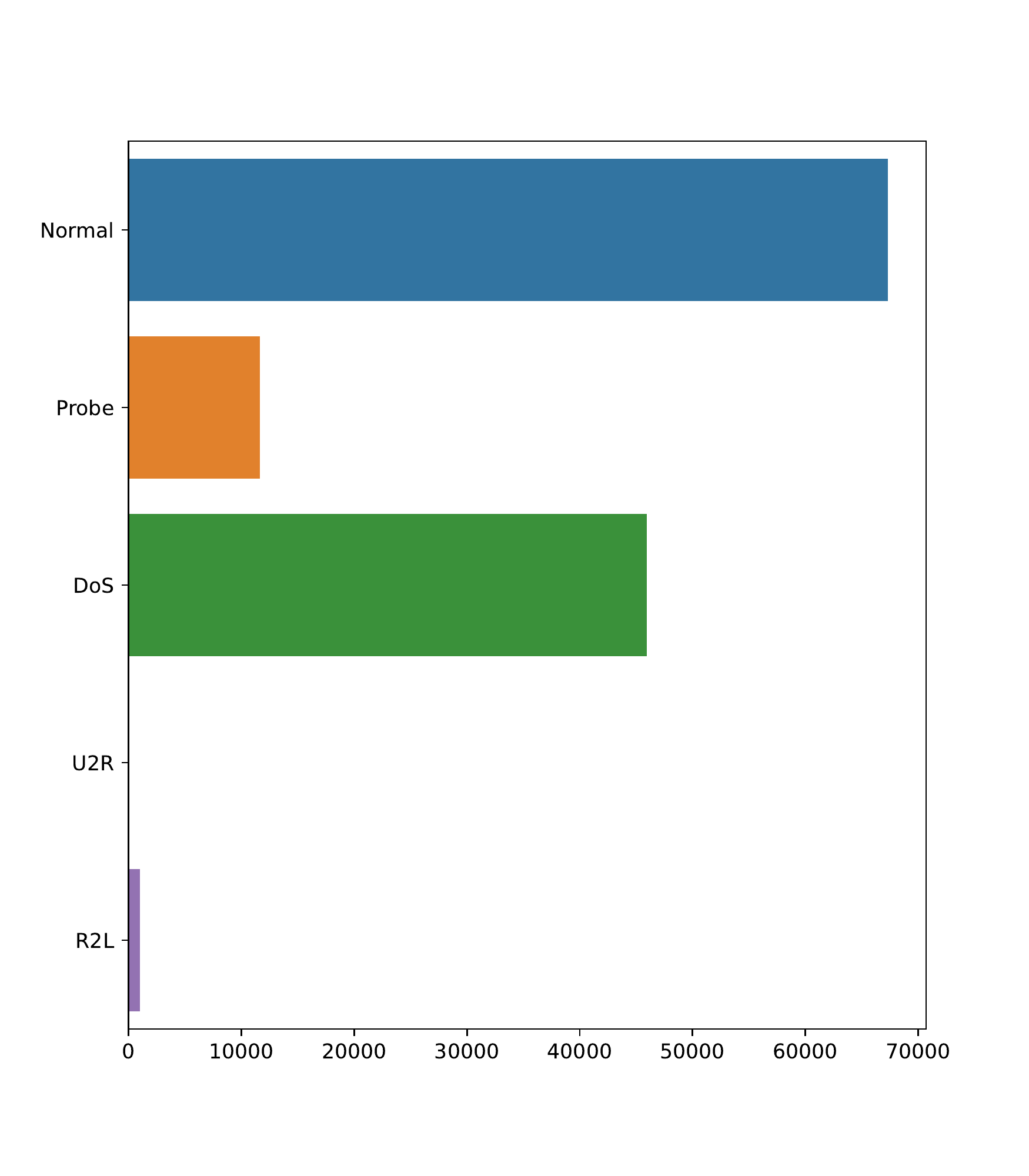}
        \caption{ORG}
    \end{subfigure}
    \hfill
    \begin{subfigure}[b]{0.30\textwidth}
        \centering
        \includegraphics[width=\linewidth]{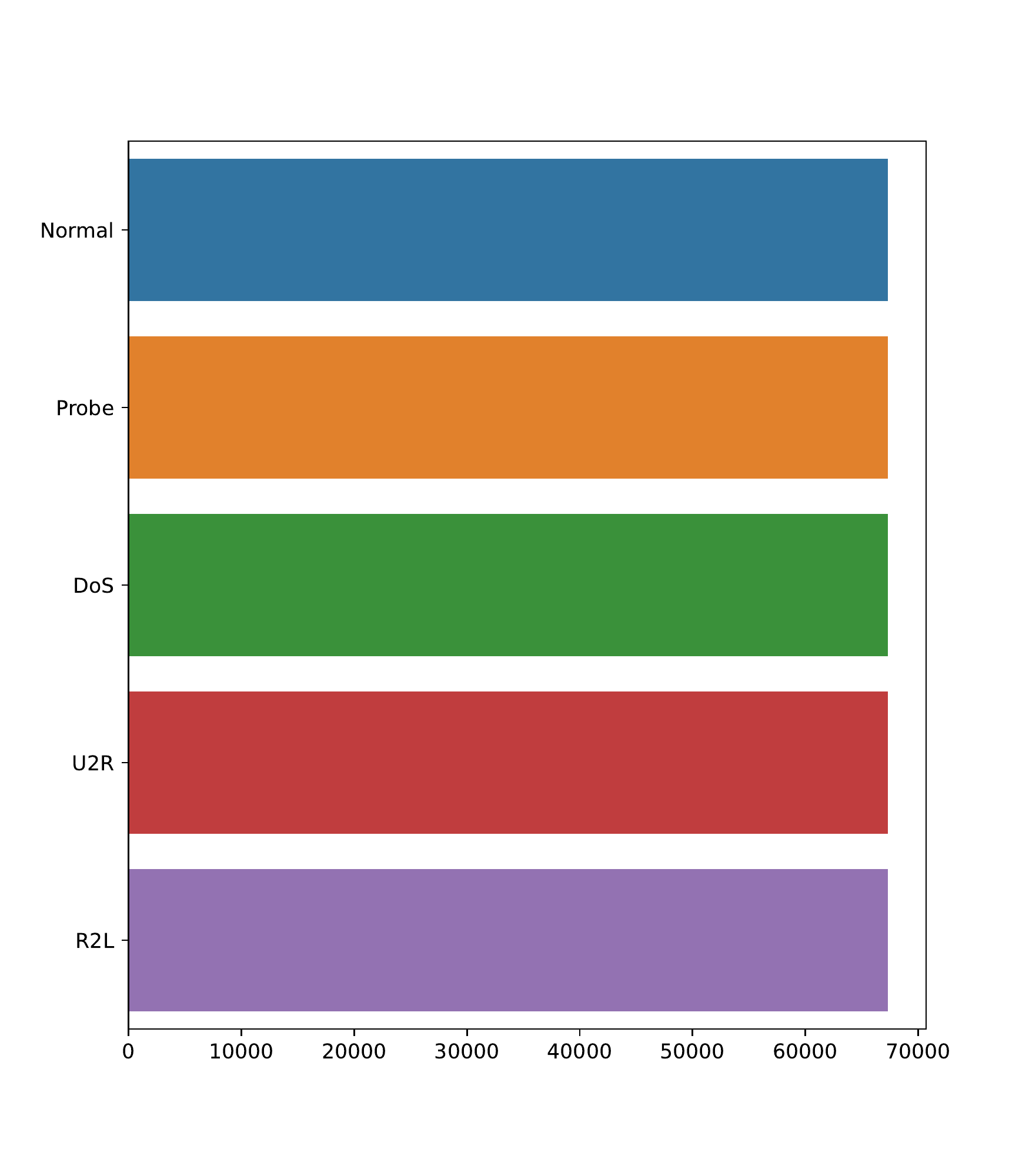}
        \caption{RndOSamp}
    \end{subfigure} 
        \hfill
    \begin{subfigure}[b]{0.30\textwidth}
        \centering
        \includegraphics[width=\linewidth]{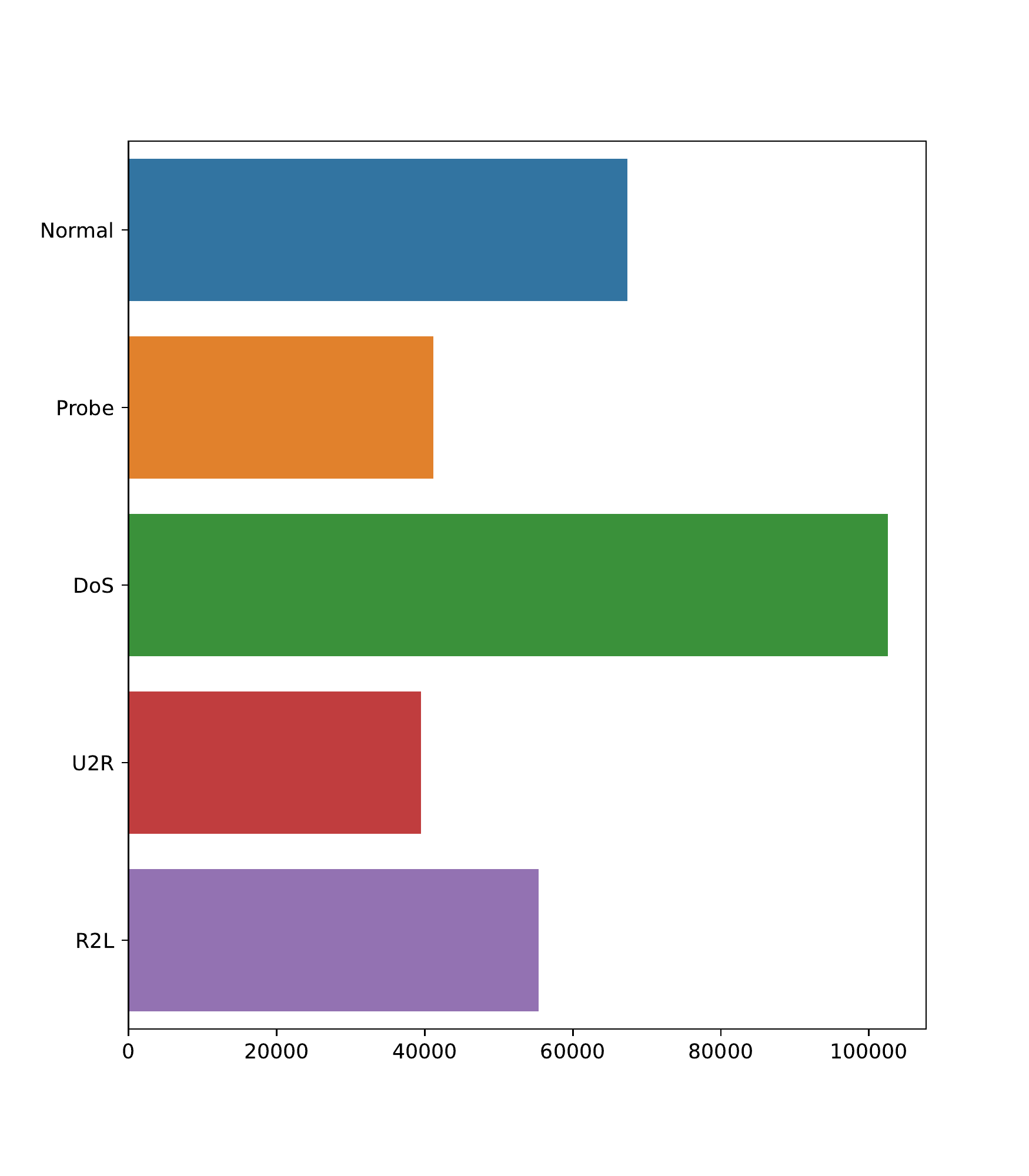}
        \caption{CTGAN}
    \end{subfigure} 
    \caption{
    Data distribution for various experimental setups.
    }
\label{fig: Data distribution}
\end{figure}

 Random oversampling is a naive technique for balancing distribution of data
 under various class types. It involves duplicating samples randomly
 from minority classes to balance the dataset. In this method, each
 member of the population stands an equal chance of being selected for
 addition to the dataset. During the entire sampling process, each subject is independently selected from the other members of the population~\citep{sharma2017pros}.
Figure~\ref{fig: Data distribution}~(b),
shows the distribution of the training samples after applying random
oversampling. Each class type has the same number of training samples. There are approximately $67000$ samples for each class type.

\subsubsection{Experimental Setup for Original Data Balanced with Synthetic Samples
  Generated using CTGAN  (CTGANSamp)}

In order to generate more realistic synthetic datasets, we use
CTGAN~\citep{xu2019modeling}. CTGAN generates synthetic data from single tabular dataset.  In the original training data, total number of samples for
normal class type is $67343$. And total number of samples for {\bf
  Probe Attack, DoS Attack, U2R Attack}, and {\bf R2L Attack} are $11656$, $45927$, $52$, and
$995$ respectively. Since the number of samples for normal class
type is already high,  we decide not to add more synthetic samples to
that class type using CTGAN.  As shown in Figure~\ref{fig: Data
  distribution}~(c),  the distribution of
samples after balancing data using synthetic samples generated by  
CTGAN are $41149$, $102589$, $39483$, and $55350$ for the attack types
{\bf  Probe Attack,  DoS Attack, U2R Attack}, and {\bf R2L Attack},
respectively. 

\subsection{Implementation Details of Various Classifiers}
Using the NSL-KDD dataset, we investigate the performance of the
following  state-of-the-art classifiers: Decision Tree (DT),
multinomial Naive Bayes (NB), Random Forest (RF), Support Vector
Machine (SVM), Feed-Forward Neural Network (FNN), Long Short term
Memory Network (LSTM), and Convolutional Neural Network (CNN). As we
mentioned earlier, we evaluated the performance of these classifiers
on (i) Original NSL-KDD dataset  (ORG), (ii) NSL-KDD dataset balanced
with random
over-sampling (RndOSamp), and (iii) NSL-KDD dataset balanced with synthetic
 data generated by CTGAN  (CTGANSamp).  

We implement  DT, NB, RF, and SVM algorithms using the
scikit-learn python package version $1.1$~\citep{scikit-learn}. We use
the Gini index splitting criteria  for 
DT and  L1 regularization for SVM. When estimating a RF, we
considered the number of trees in the forest to be $100$. With respect
to NB, we use alpha $1.0$, as smoothing parameter. 

The following three neural networks have been used: FNN, LSTM, and
CNN. With FNN, we use  three hidden layers, each containing $50$, $30$, and $20$ neurons respectively; and the output layer had
five neurons. In the final layer, we use softmax function to do the
final classification. We use relu as activation function, adam as 
optimizer, and categorical cross 
entropy as a loss function. In total this network has $5285$ trainable
parameters. We design a two-layer LSTM. Each layer in the LSTM had $100$
units. The activation function, optimizer, loss function and final
layer are same as in FNN. Since, our dataset consists of
one-dimensional sequence of data,  we used a single one dimensional
CNN (conv1D). We used $32$ filters with kernel size of $3$. We used
maxpooling with pool size of $2$. Next, we use a dense hidden layer
with 100 neurons and the final layer had five neurons. Like FNN and
LSTM,  we used the same activation function, optimizer, and loss
function for CNN. Each of these networks has been trained for $100$
epochs with early stopping. We implement all three neural networks using tensorflow and
Nvidia GPU driver version $455.32.00 $ with cuda version $11.1 $.

\subsection{Results}

The NSL-KDD dataset comes with two testing datasets (KDDTest+ and KDDTEST21-).
Table \ref{tab:performance on KDDtest+} shows the performance
of various state-of-the-art classifiers tested on KDDTest+ with ORG,
RndOSamp, and CTGANSamp.  It presents
weighted-average scores \citep{patro2014augmenting,} of accuracy, precision, recall, and $F_{1}$
score.

\begin{table}[t!]  
\caption{Performance of different ML models on KDDTest+ test set where models were trained with ORD, RndOSamp, and CTGAN}
\begin{center}
\fontsize{8pt}{14pt}
\selectfont	
\begin{tabular}{c|cccc|cccc|cccc}
\hline
\textbf{Classifier}&\multicolumn{4}{|c|}{\textbf{ORG}} & \multicolumn{4}{|c|}{\textbf{RndOSamp}} & \multicolumn{4}{|c}{\textbf{CTGANSamp}}\\
 
\textbf{Name} & \textbf{\textit{Acc}}& \textbf{\textit{Pre}}& \textbf{\textit{Rec}} & \textbf{\textit{$F_{1}$}}& \textbf{\textit{Acc}}& \textbf{\textit{Pre}}& \textbf{\textit{Rec}}& \textbf{\textit{$F_{1}$}}
&\textbf{\textit{Acc}}& \textbf{\textit{Pre}}& \textbf{\textit{Rec}}& \textbf{\textit{$F_{1}$}}\\
 \hline

 DT & $0.7315 $ & $0.7144$ & $0.7315$ & $0.6830$ & $ 0.7458 $ & $ 0.7992$ & $  0.7458$ & $ 0.7036$ & $\bf{0.7522}$& $\bf{0.7995}$ & $ \bf{0.7522} $ & $ \bf {0.7203}$ \\

 SVM &  $ 0.7014 $ & $0.6547$ & $0.7014$ & $0.6561$ & $0.6935$ & $ \bf 0.7169 $ & $0.6935$ & $\bf 0.6964$ & $\bf{0.7326 }$& $0.6869$ & $\bf 0.7326$ & $0.6842 $\\

 RF & $0.7393$ & $\bf 0.8162$ & $ 0.7393$ & $\bf 0.6917$  & $0.7355$ & $0.7485 $ & $0.7355$ & $ 0.6876$ 
 & $\bf{0.7394 }$& $\bf 0.8162$ & $\bf 0.7394$ & $0.6916$\\

 NB & $ 0.6105 $ & $ 0.5395 $ & $0.6105$ & $0.5254$ & $0.4483$ & $\bf 0.5695$ & $ 0.4483$ & $ 0.4937$ & $\bf{0.6273}$& $0.5168$ & $\bf0.6273$ & $\bf0.5592$\\

 FNN & $0.7534 $ & $0.7304 $ & $0.7534$ & $0.7200$ & $0.7587  $ & $0.7567$ & $0.7587$ & $\bf 0.7399$ &$ \bf{0.7736 }$&$\bf 0.8081$& $\bf 0.7736$ & $0.7372$\\

 LSTM & $ 0.7629 $ & $0.8010  $ & $0.7629$ & $ 0.7260$ & $0.7498 $ & $0.7899$ & $0.7498$ & $0.7130 $ &$ \bf{0.7762} $& $\bf 0.8182 $ & $\bf 0.7762$ & $\bf0.7462$\\

 CNN & $0.7505$ & $0.6887$ & $0.7505 $ & $0.7021$ & $0.7517$ & $0.7849$ & $0.7517$ & $0.7156$ & $\bf{0.7717}$ & $\bf0.8037$ & $ \bf0.7717$ & $\bf0.7344$ \\
 \hline

\end{tabular}
\label{tab:performance on KDDtest+}
\end{center}
\end{table}

\begin{table}[t!]
\caption{Performance of different ML models on KDDTest21- test set where models were trained with ORD, RndOSamp, and CTGANSamp}
\begin{center}
	\fontsize{8pt}{14pt}
	\selectfont
\begin{tabular}{c|cccc|cccc|cccc}
\hline
\textbf{Classifiers}&\multicolumn{4}{|c|}{\textbf{ORG}} & \multicolumn{4}{|c|}{\textbf{RndOSamp}} & \multicolumn{4}{|c}{\textbf{CTGANSamp}}\\
 
\textbf{Name} & \textbf{\textit{Acc}}& \textbf{\textit{Pre}}& \textbf{\textit{Rec}} & \textbf{\textit{$F_{1}$}}& \textbf{\textit{Acc}}& \textbf{\textit{Pre}}& \textbf{\textit{Rec}}& \textbf{\textit{$F_{1}$}}
&\textbf{\textit{Acc}}& \textbf{\textit{Pre}}& \textbf{\textit{Rec}}& \textbf{\textit{$F_{1}$}}\\
 \hline
 
 DT & $ 0.4917$ & $ 0.5956 $ & $ 0.4917 $ & $ 0.4573$ & $ 0.5248$ & $0.7344 $ & $ 0.5248 $ & $0.5020$ & $\bf{0.5423 }$& $ \bf 0.7560$ & $ \bf 0.5423$ & $ \bf 0.5345$ \\

 SVM &  $  0.4349 $ & $0.5120 $ & $0.4349$ & $0.4248 $ & $ 0.4823$ & $0.5585$ & $0.4823$ & $\bf 0.4910$ & $\bf{0.4993}$& $\bf0.5664$ & $\bf0.4993$ & $0.4761$\\

 RF &  $0.5036$ & $ 0.8085 $ & $ 0.5036$ & $0.4846$ & $0.4963 $ & $0.6804$ & $0.4963$ & $0.4772$ & $\bf{0.5043}$& $\bf0.8092$ & $\bf0.5043$ & $\bf0.4855$\\

 NB & $0.2659$ & $\bf 0.3144 $ & $0.2659 $ & $0.2023$ & $0.2772$ & $0.3006$ & $0.2772$ & $\bf 0.2787$ & $\bf{0.3503}$& $0.2402$ & $\bf 0.3503$ & $0.2699$\\

 FNN & $ 0.5344$ & $0.5975 $ & $0.5344 $ & $0.5148$ & $0.5464 $ & $0.6425$ & $0.5464$ & $0.5548 $ &$ \bf{0.5719 }$&$\bf  0.7656$& $ \bf 0.5719 $ & $  \bf 0.5569$\\

 LSTM & $ 0.5535 $ & $ 0.7528 $ & $0.5535$ & $ 0.5370$ & $0.5279 $ & $0.7265$ & $0.5279   $ & $ 0.5088$ &$ \bf{0.5774 } $& $\bf 0.7837$ & $\bf 0.5774$ & $\bf0.5752$\\

 CNN & $0.5289$ & $0.5558$ & $0.5289$ & $0.4997$ & $0.5315$ & $0.7352$ & $0.5315$ & $0.5236$ &$ \bf{0.5661}$ & $ \bf0.7575$ & $\bf0.5661$ & $\bf0.5531$ \\
 \hline

\end{tabular}
\label{tab:performance on KDDtest21-}
\end{center}
\end{table}

Table~\ref{tab:performance on KDDtest21-} shows the performance of
various classifiers on KDDTest21- with ORG, RndOSamp, and
CTGANSamp. 
Performance of various classifiers,   trained on dataset balanced with synthetic data
generated using CTGAN (CTGANSamp),  on both test
datasets (KDDTest21- and KDDTest21+)  shows an improvement in accuracy
ranging from $1\%$ to $8\%$. 
We notice that DT, LSTM, and CNN classifiers trained on CTGANStamp consistently outperformed their counterparts trained on ORG and RndOSamp
in both KDDTest+ and KDDTest21- datasets for all the quantitative metrics: accuracy, precision, recall, and the $F_{1}$ scores.
Similarly, we observe significant improvements in \emph{recall} for all ML classifiers for CTGANSamp compared to ORG and RndOSamp. Using CTGAN to balance the original data, it is evident that the $FN$ number is low and the $TP$ rate is high. 
It is also important to highlight that the accuracy of all the classifiers is consistently better when they were trained on CTGANStamp compared to when they were trained using ORG and RndOSamp. When all the classifiers are compared for $F_{1}$ score, it can be observed that all the classifiers either show better or competitive (e.g., within 1.5 percentage point) performance when trained employing CTGANStamp versus ORG and RndOSamp. However, the precision for CTGANSamp is not as good as it is for ORG or RndOSamp for SVM and NB classifiers. Additionally, the experimental results we obtain also show that
accuracy  of some classifiers decrease under RndOSamp compared to
their accuracy under ORG. Overall, it can be conclusively claimed that all the classifiers show significant improvement for various metrics on both datasets. Under CTGANSamp, in the following, we further discuss quantitative results

As shown in Table~\ref{tab:performance on
  KDDtest+},  accuracy of DT for ORG is $73.15\%$; it
increases around $1\%$ under  RndOSamp (i.e., $74.58\%$); it further increases
to   $75.22\%$ under CTGANSamp. For the dataset KDDTest21-, we can observe similar trend. Accuracy of DT for ORG,
RndOSamp, and CTGANSamp are $49.17\%$, $52.48\%$, and $54.23\%$, respectively,
as shown in Table~\ref{tab:performance on KDDtest21-}.
For KDDTest+, CTGANSamp's $F_{1}$ scores for the classifiers DT, NB, LSTM, and CNN are significantly higher than the competing methods ORG and RndOSamp. For instance, the classifier NB achieved $55.92\%$ $F_{1}$ score for CTGANSamp, while ORG and RndOSamp achieved $52.54\%$ and $49.37\%$, respectively. Table \ref{tab:performance on KDDtest+}, shows, in the case of some classification methods such as SVM, RF, and FNN, ORG and RndOSamp achieved the highest $F_{1}$ scores. However, they differ slightly from CTGANSamp's $F_{1}$ scores. As we can see, the difference between ORG and CTGANSamp for RF is $0.01\%$. Table \ref{tab:performance on KDDtest21-} shows that there are significant improvements in CTGANSamp $F_{1}$ scores for DT, RF, FNN, LSTM, and CNN compared with ORG and RndOSamp. In the same dataset,  for SVM and NB, RndOSamp achieved the highest $F_{1}$ scores. Again, there is a slight difference between CTGANSamp and RndOSamp which is around $1\%$. Table \ref{tab:performance on KDDtest+} indicates that the accuracy of NB for RndOSamp on KDDTest+ decreases by $16\%$ compared to ORG. The performance of the NB classifier is determined by the distribution of the training dataset. When duplicate samples are used to balance the training samples and change the distribution of the true dataset, the NB become more biased to some identical samples, therefore, achieved lower accuracy upon testing. Several other classifiers in Table \ref{tab:performance on KDDtest+} also decrease the overall accuracy, including SVM, RF, and LSTM for RndOSamp. As shown in Table \ref{tab:performance on KDDtest21-}, the overall accuracy for RF and LSTM decrease to $0.73\%$ and $2.56\%$, respectively, for RndOSamp.

\begin{figure}[t]
    \centering
    \begin{subfigure}[b]{0.22\textwidth}
        \centering
        \includegraphics[width=\linewidth]{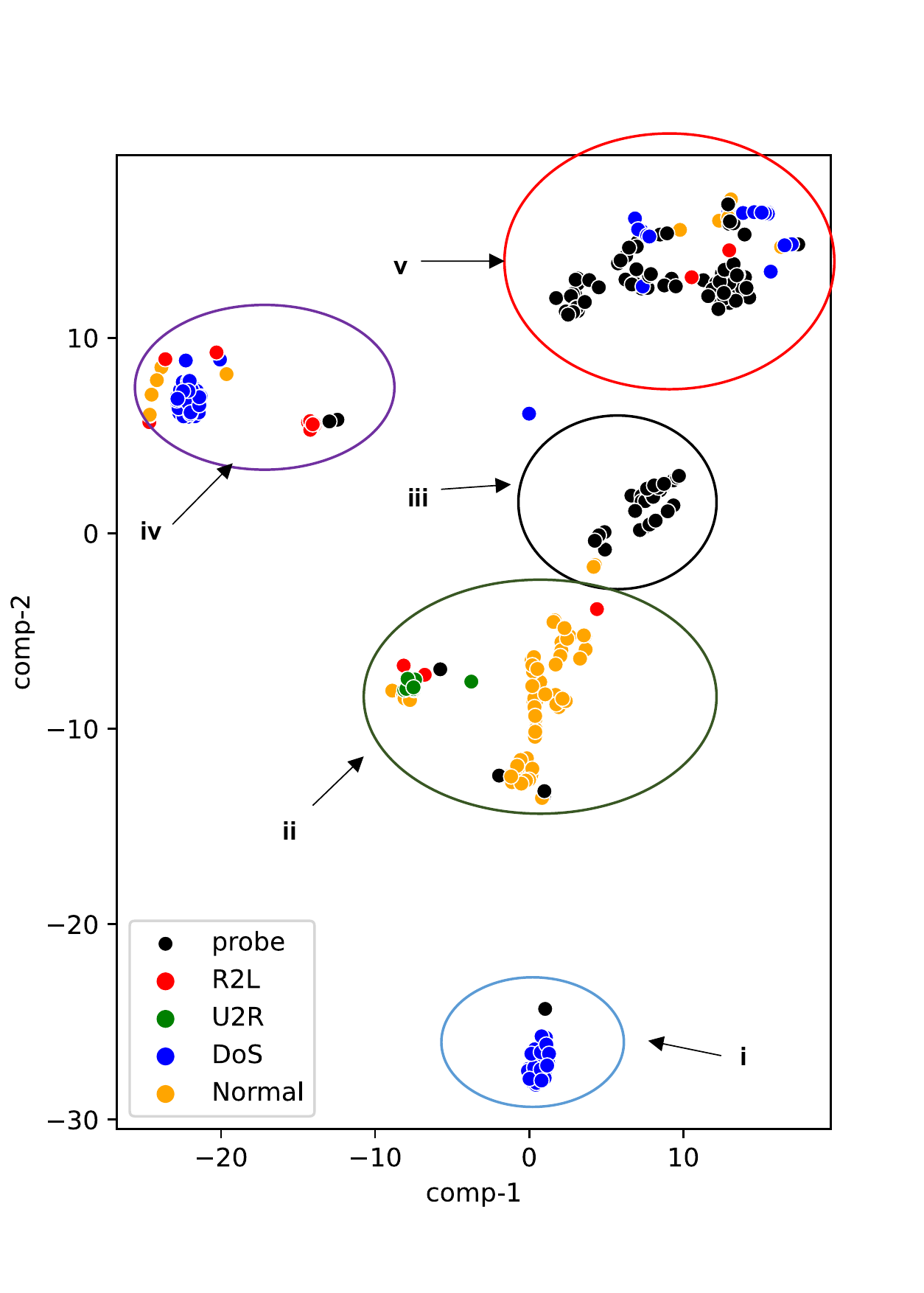}
        \caption{Ground Truth}
    \end{subfigure}
    \hfill
    \begin{subfigure}[b]{0.22\textwidth}
        \centering
        \includegraphics[width=\linewidth]{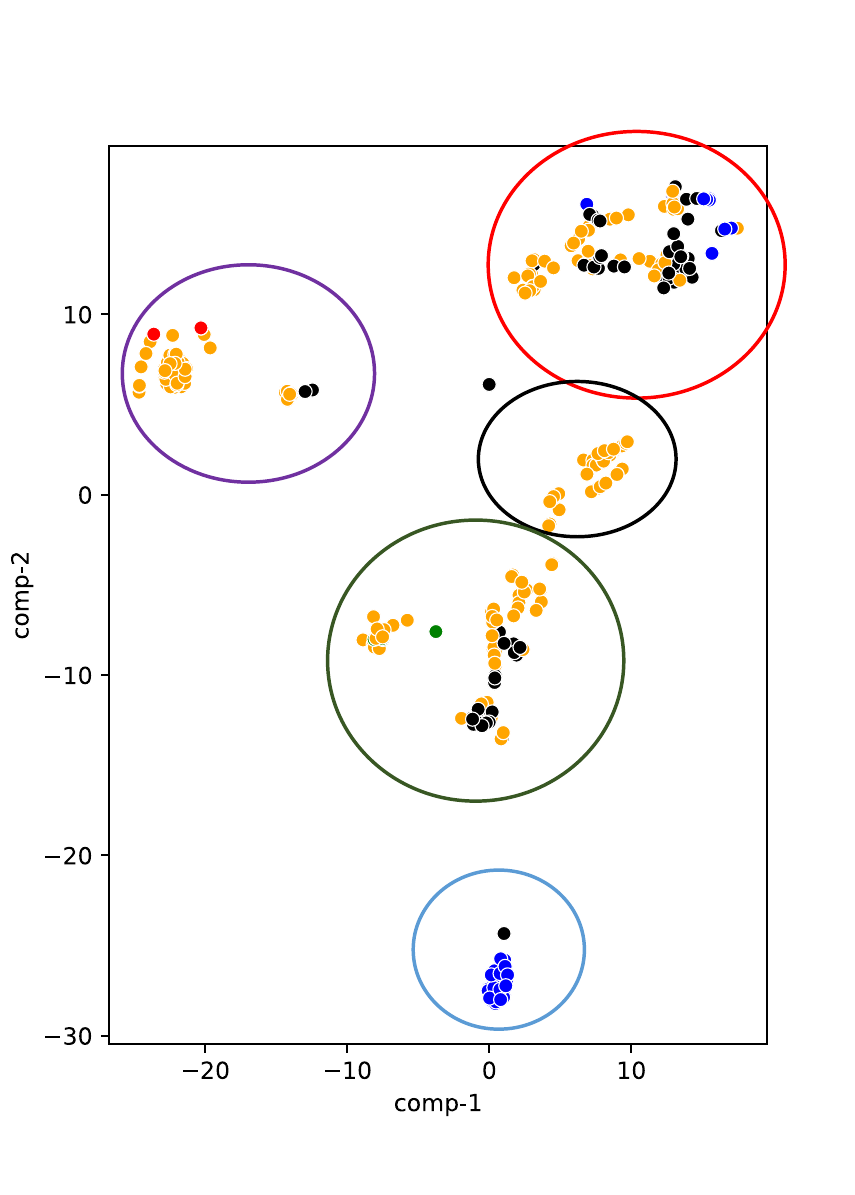}
        \caption{ORG}
    \end{subfigure} 
        \hfill
    \begin{subfigure}[b]{0.22\textwidth}
        \centering
        \includegraphics[width=\linewidth]{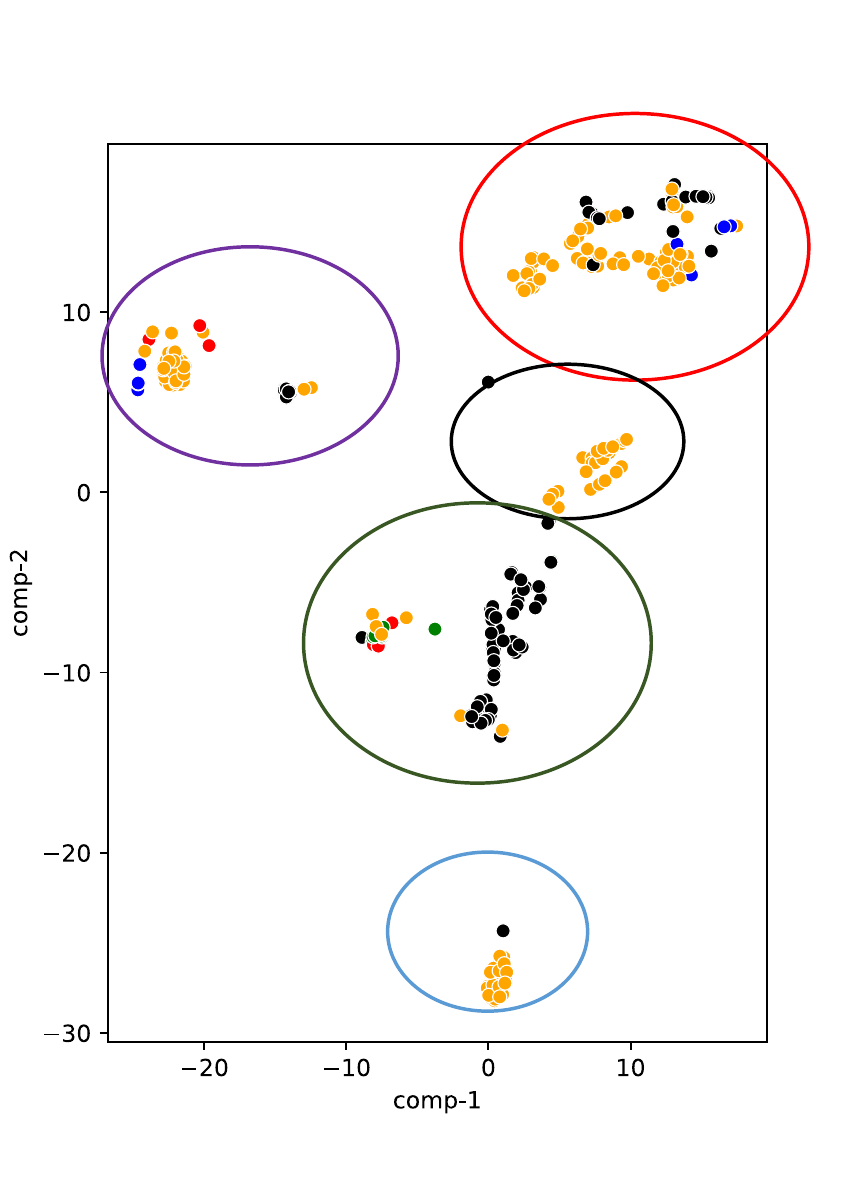}
        \caption{RndOSamp}
    \end{subfigure} 
    \hfill
        \begin{subfigure}[b]{0.22\textwidth}
        \centering
        \includegraphics[width=\linewidth]{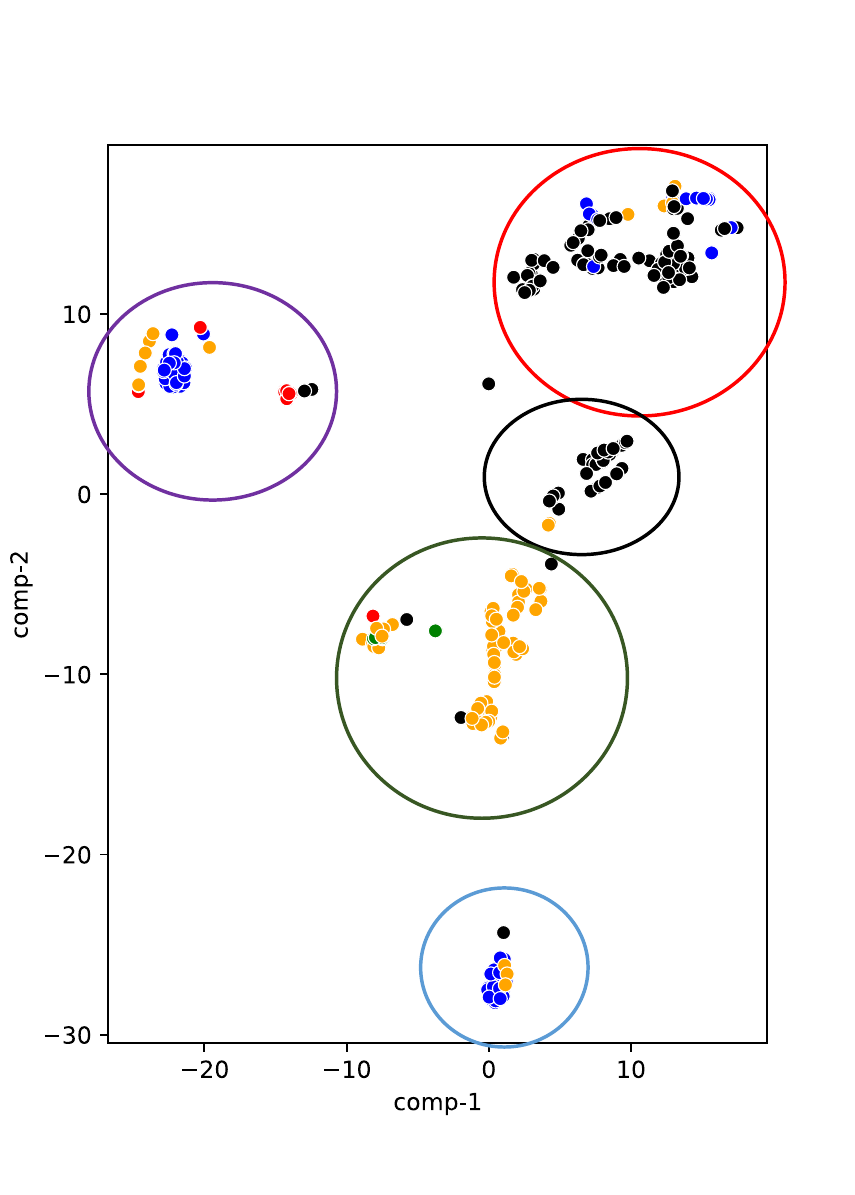}
        \caption{CTGANSamp}
    \end{subfigure}
    \caption{ Qualitative analysis of the
  CNN classifier trained with  ORG, RndOSamp, and CTGANSamp.}
\label{fig: t-sne selested samples}
\end{figure}

We randomly drew $300$  samples from KDDTest+ and perform
t-SNE projection on the selected data samples for qualitative analysis. Figure~\ref{fig: t-sne
  selested samples} shows the t-SNE visualization for all three
experiments alongside ground truth for the classifier CNN. Figure \ref{fig: t-sne selested samples}(a) shows the actual classes, followed by ORG's classification in \ref{fig: t-sne selested samples}(b), then RndOSamp's classification in \ref{fig: t-sne selested samples}(c), and then CTGANSamp's classification in \ref{fig: t-sne selested samples}(d).
In Figure \ref{fig: t-sne selested samples}, we mark five circles (i.e., classes or clusters) shown in  
blue~(I), green~(II), black~(III), purple~(IV), and red~(V) colors. We
can see that the classification based on CTGANSamp is much
closer to the actual truth. 
The blue circle in the ground truth indicates that the majority of the samples are of the DoS class type. In the same circle, ORG performs very well and is in line with the truth. The RndOSamp, on the other hand, predicts the opposite of the truth. CTGAN has incorrectly predicted some DoS samples as normal in the same circle, but overall it is able to detect most of the DoS samples. The majority of the samples in the green circle of ground truth are normal, and only a few are U2R, R2L, and probe samples. ORG is not able to detect any of the U2R and R2L samples, and some of the normal samples are also labeled as probes. Among all the experiments on this circle, RndOSamp performs the worst. It label many normal samples as probes. CTGANSamp, on the other hand, detects all the normal samples along with the minority classes U2R and R2L. In the black circle of the ground truth, the majority of the samples are probes, where ORG and RndOSamp predict all the samples to be normal. CTGAN's prediction, on the other hand, is similar to the ground truth. It is also the case with the purple and red circles, where CTGANSamp's prediction results are the same as the actual results. In contrast, both ORG and RndOSamp fail to capture the actual truth for the same circles.

Finally, we also perform statistical significant test on KDDTest+, Student's T-test, on
these three experiments to check if there is any significant
difference and P-values are presented in table \ref{tab:t-test KDDTest+}. We notice that the P-values are consistently very small
when we compare the performance of CTGANSamp with ORG and RndOSamp. On
the other hand, P-values for ORG and RndOSamp are not always small
(for example, the P-value of RF is $0.91$). This result also signifies that ML classifiers show improvement when they were trained using dataset augmented with CTGAN.

\begin{table}[htbp]
\caption{T-Test on performance of various classifiers trained with
  ORG, RndOSamp, and CTGANSamp}
\begin{center}
\begin{tabular}{c|c|c|c|c|c|c|c}
\hline
\textbf{Classifiers}& {\textbf{DT}}&{\textbf{SVM}}&{\textbf{RF}}&{\textbf{NB}}& {\textbf{FNN}}&{\textbf{LSTM}}&{\textbf{CNN}}\\

\textbf{Comparing approaches} & \textbf{\textit{P-value}}& \textbf{\textit{P-value}}& \textbf{\textit{P-values}}& \textbf{\textit{P-value}}&\textbf{\textit{P-value}}& \textbf{\textit{P-value}}& \textbf{\textit{P-value}}\\

\hline
 ORG, CTGANSamp & $3.10 e^{-10}$ &$3.34 e^{-78}$& $2.52 e^{-16}$& $3.01 e^{-253}$&$0.05$&$0.001$& $1.64 e^{-37}$\\
 ORG, RndOSamp &  $2.64 e^{-17}$&$1.73 e^{-301}$& $0.91 $& $0.0 $& $1.79 e^{-16}$&$6.65 e^{-13}$& $2.05 e^{-18}$\\
 RndOSamp, CTGANSamp & $2.69 e^{-12}$ &$5.70 e^{-128}$&$9.99 e^{-17}$&$7.94 e^{-151}$& $3.46 e^{-26}$& $8.01 e^{-05}$ & $0.00028$\\

 \hline

\end{tabular}
\label{tab:t-test KDDTest+}
\end{center}
\end{table}

\section{Related Works}\label{related-work}

In this section, we first discuss some
intrusion detection techniques based on Machine Learning, presented in
the literature. Then, we discuss data augmentation techniques used in
various applications.

\subsection{Machine Learning based  Intrusion Detection}

Machine Learning has been used extensively in designing and
implementing IDSes.  
Ever {\it et al.}~\citep{Ever-2019} used three machine learning models,
ANN, SVM, and DT in their study. The primary  goal of this study was
to determine the optimal machine learning technique. 
As part of their experiments, they used 60\% and 70\% of
the dataset NSL-KDD for training, and the rest of the dataset for
testing. Based on their experiments, they showed that DT  achieved better
accuracy compared to the other two.

A new approach to detect intrusion in computer networks was introduced
by {\it et al.}~\citep{XianweiGao-2019}.  In order to address the data
imbalance problem in NSL-KDD dataset, they proposed a MultiTree
algorithm using DT of four levels, with the proportions of the types
of classes adjusted accordingly. The authors introduced a model in
which they ensembled DT, RF, K-NN, and DNN and used their adaptive
voting algorithm to decide on classification. 

To build effective IDSes, in depth analysis of
network data is mandatory, as the volume of network data
increases. Due to the different types of protocols used on the
Internet, we have diverse network data. Therefore, it is difficult to
distinguish between normal network traffic and attack traffic. Shone {\it et al.}~\citep{Shone-2018} studied the feasibility and
sustainability of current approaches in network intrusion
detection.  Deep
and shallow learning were combined in their model. For unsupervised
feature learning, the authors applied two layers of non-symmetric deep
auto-encoders (NDAE). Unlike conventional auto encoders, the NDAE
contains no decoder. In order to perform the final classification of
the network traffic into normal and attack, RF was used. Based on
NSL-KDD and KDD99 datasets, the authors evaluated their model using
five and thirteen layers of classification. To overcome the problem of
over-fitting and under-fitting, they performed a 10-fold cross
validation. Due to the imbalanced nature of the datasets, the false
alarm rate was high in some attack classes.

Yin {\it et
  al.}~\citep{yin2017deep} presented a two-step approach for intrusion
detection based on deep learning. One hot encoding was used to
transform categorical data to numerical values during the
preprocessing stage. In the following step, min-max method was used to
normalize the dataset due to large variations in the data
distribution. To classify data, recurrent neural networks~(RNNs) 
with forward propagation and backward propagation were used. In the
forward propagation method, output values were calculated, and the
backward propagation method calculated the error and updated the
weights. Cross-entropy was used to compute the difference between the
output values produced by forward propagation and the true
value. Using this methodology, both binary and multiclass
classification were performed.

Javaid {\it et al.}~\citep{javaid2016deep} introduced a deep learning
 technique based on Auto Encoder~(AE) for feature representation and feature
 learning. They used softmax regression for
 classification. Additionally, in the preprocessing stage, they
 transformed categorical features into continuous features and
 normalized the whole dataset using min-max method. They performed two
 types of evaluations. In order to do cross validation, they used
 training data for both training and testing. In the second approach,
 they used different datasets for testing and training. 
 
 In all of the above works, NSL-KDD dataset was used. As we saw,  this
 dataset is imbalanced. However, none of the authors addressed this
 issue. The purpose of our study is to focus on the data imbalance
 problem and to investigate how this impacts the overall performance
 of various  machine learning models.

\subsection{Augmentation Techniques Applied to Various Applications} 
Synthetic data generation or data augmentation has been used in a
variety of applications such as image classification, natural language
processing. Various augmentation techniques have been
proposed, primarily based on deep learning models. In this subsection,
we review some recent works  on data augmentation and how this
technique was applied in different areas of research.  

 Shorten {\it et al.}~\citep{shorten2019survey} presented a critical
 survey on image data augmentation using deep learning
 techniques. They explored the use of data augmentation, a data-space
 approach to the problem of limited data. Additionally, they state
 that data augmentation encompasses a suite of techniques  to augment
 the size and quality of training datasets in order to build better
 deep learning models. This survey discussed image augmentation
 algorithms including geometric transformations, color space
 augmentations, kernel filters, mixing images, random erasing, feature
 space augmentation, adversarial training, generative adversarial
 networks, neural style transfer, and meta-learning. A significant
 portion of the survey is devoted to the application of GANs for augmentation.

 Li {\it et al.}~\citep{li2020intelligent} proposed a novel deep
 learning technique for rotating machinery fault diagnosis.
 Generally, the following five data augmentation techniques were examined:
 additional Gaussian noise, masking noise, signal translation,
 amplitude shifting, and time stretching. Sample-based as well as
 dataset-based augmentation techniques were considered. They used two
 datasets to conduct their experiments, namely: Bearing Data
 Center of Case Western Reserve University~(CWRU) and Intelligent
 Maintenance System~(IMS). Their  approach was able to achieve $99.9\%$
 accuracy.

Zhou {\it et al.}~\citep{zhou2020forecasting} proposed a novel approach combining data augmentation and deep learning methods, which addresses the issue of a lack of training samples in deep learning when used to forecast emerging technologies. In order to construct a sample dataset, Gartner's hype cycle and multiple patent features were utilized. As a second step, a generative adversarial network was used to create many synthetic samples (i.e., data augmentation) in order to expand the sample dataset. Lastly, a deep neural network classifier was trained with the augmented data set to forecast emerging technologies, and it was able to accurately predict up to $77\%$ of the emerging technologies in a given year. Based on patent data from 2000-2016, this approach was used to predict emerging technologies in Gartner's hype cycles for 2017. A total of four out of six emerging technologies were accurately predicted, demonstrating the precision and accuracy of the proposed method. This article showed that deep learning now can be used to forecast emerging technologies with limited training samples.
 
ML-classifiers trained with imbalanced datasets affect their
performance.
We utilized synthetic data generated with CTGAN, to augment and
balance a well known  training dataset to study its effect on the
performance of various well-known  ML-classifiers.

\section{Conclusion}\label{conclusion}
Over the past several years, many researchers used Machine Learning in
designing and implementing IDSes. They used different datasets for
training ML classifiers. Some of the datasets used in such works are:
NSL-KDD~\citep{NSL-KDDdataset},
UNSW-NB15~\citep{moustafa2015UNSW,moustafa2016evaluation}, CICIDS
2017~\citep{sharafaldin2018toward}, 
 In many of the datasets used for designing IDSes, data are
imbalanced (i.e., not all classes have equal amount of data). With
unbalanced data, the predictive models developed using ML  algorithms
may produce unsatisfactory classifiers which would affect accuracy in
predicting intrusions. Traditionally, researchers used over-sampling
and under-sampling techniques to balance data in datasets. In this
work, we use over-sampling, and also use  a synthetic data
generation method, called  Conditional Generative Adversarial Network
(CTGAN) to balance data and study their  effect on various ML
classifiers. To the best of our knowledge,  no one else has used CTGAN
to generate synthetic samples to balance datasets designed for
intrusion detection in computer networks. Based on extensive
experiments with the widely used  dataset NSL-KDD,  we found that
training ML models on  data balanced
with  synthetic samples generated by CTGAN
increased prediction accuracy by as much as  $8\%$,  compared to training
the same ML models over unbalanced data. Our experiments also show
that the accuracy of   some ML models trained over data balanced with
random over-sampling declined compared to the same ML models trained over
unbalanced data.


\begin{thebibliography}{10}
\expandafter\ifx\csname url\endcsname\relax
  \def\url#1{\texttt{#1}}\fi
\expandafter\ifx\csname urlprefix\endcsname\relax\def\urlprefix{URL }\fi
\expandafter\ifx\csname href\endcsname\relax
  \def\href#1#2{#2} \def\path#1{#1}\fi

\bibitem{DINA2021100462}
A.~S. Dina, D.~Manivannan,
  \href{https://www.sciencedirect.com/science/article/pii/S2542660521001037}{Intrusion
  detection based on machine learning techniques in computer networks},
  Internet of Things 16 (2021) 100462.
\newblock \href {https://doi.org/https://doi.org/10.1016/j.iot.2021.100462}
  {\path{doi:https://doi.org/10.1016/j.iot.2021.100462}}.
\newline\urlprefix\url{https://www.sciencedirect.com/science/article/pii/S2542660521001037}

\bibitem{lastlinecostanalysis}
link, Lastline article available online:
  https://www.lastline.com/blog/examine-tco-of-a-of-a-network-intrusion-detection-system/.

\bibitem{khraisat2019survey}
A.~Khraisat, I.~Gondal, P.~Vamplew, J.~Kamruzzaman, Survey of intrusion
  detection systems: techniques, datasets and challenges, Cybersecurity 2~(1)
  (2019) 1--22.

\bibitem{Anderson-1972}
J.~P. Anderson, Computer security technology planning study. volume 2. (1972).

\bibitem{Bridges-2020}
R.~Bridges, T.~Glass-Vanderlan, M.~Iannacone, M.~Vincent, Q.~Chen, A survey of
  intrusion detection systems leveraging host data, ACM computing surveys
  52~(6) (2020) 1--35.

\bibitem{Stallings-2018}
W.~Stallings, L.~Brown, Computer Security Principles and Practice (4th
  edition), Pearson, ISBN-13: 978-0-13-479410-5, 2018.

\bibitem{Buczak-2016}
A.~L. Buczak, E.~Guven, A survey of data mining and machine learning methods
  for cyber security intrusion detection, IEEE Communications Surveys \&
  Tutorials 18~(2) (2016) 1153 -- 1176.
\newblock \href {https://doi.org/10.1109/COMST.2015.2494502}
  {\path{doi:10.1109/COMST.2015.2494502}}.

\bibitem{saranya2020performance}
T.~Saranya, S.~Sridevi, C.~Deisy, T.~D. Chung, M.~A. Khan, Performance analysis
  of machine learning algorithms in intrusion detection system: A review,
  Procedia Computer Science 171 (2020) 1251--1260.

\bibitem{buczak2015survey}
A.~L. Buczak, E.~Guven, A survey of data mining and machine learning methods
  for cyber security intrusion detection, IEEE Communications surveys \&
  tutorials 18~(2) (2015) 1153--1176.

\bibitem{abd2013review}
S.~M. Abd~Elrahman, A.~Abraham, A review of class imbalance problem, Journal of
  Network and Innovative Computing 1~(2013) (2013) 332--340.

\bibitem{chawla2004special}
N.~V. Chawla, N.~Japkowicz, A.~Kotcz, Special issue on learning from imbalanced
  data sets, ACM SIGKDD explorations newsletter 6~(1) (2004) 1--6.

\bibitem{xu2019modeling}
L.~Xu, M.~Skoularidou, A.~Cuesta-Infante, K.~Veeramachaneni, Modeling tabular
  data using conditional {GAN}, arXiv preprint arXiv:1907.00503 (2019).

\bibitem{hasan2019attack}
M.~Hasan, M.~M. Islam, M.~I.~I. Zarif, M.~Hashem, Attack and anomaly detection
  in iot sensors in iot sites using machine learning approaches, Internet of
  Things 7 (2019) 100059.

\bibitem{mottini2016relative}
A.~Mottini, R.~Acuna-Agost, Relative label encoding for the prediction of
  airline passenger nationality, in: 2016 IEEE 16th International Conference on
  Data Mining Workshops (ICDMW), IEEE, 2016, pp. 671--676.

\bibitem{wang2012comparative}
X.~Wang, L.~Wang, Y.~Qiao, A comparative study of encoding, pooling and
  normalization methods for action recognition, in: Asian Conference on
  Computer Vision, Springer, 2012, pp. 572--585.

\bibitem{svensen2007pattern}
M.~Svens{\'e}n, C.~M. Bishop, Pattern recognition and machine learning (2007).

\bibitem{aich2018nonlinear}
S.~Aich, K.~Younga, K.~L. Hui, A.~A. Al-Absi, M.~Sain, A nonlinear decision
  tree based classification approach to predict the parkinson's disease using
  different feature sets of voice data, in: 2018 20th International Conference
  on Advanced Communication Technology (ICACT), IEEE, 2018, pp. 638--642.

\bibitem{syamala2020filter}
M.~Syamala, N.~J. Nalini, A filter based improved decision tree sentiment
  classification model for real-time amazon product review data, International
  Journal of Intelligent Engineering and Systems 13~(1) (2020) 191--202.

\bibitem{reges2021decision}
O.~Reges, A.~E. Krefman, S.~T. Hardy, Y.~Yano, P.~Muntner, D.~M. Lloyd-Jones,
  N.~B. Allen, Decision tree-based classification for maintaining normal blood
  pressure throughout early adulthood and middle age: Findings from the
  coronary artery risk development in young adults (cardia) study, American
  journal of hypertension 34~(10) (2021) 1037--1041.

\bibitem{kim2021decision}
S.-H. Kim, I.-J. Moon, S.-H. Won, H.-W. Kang, S.~K. Kang, Decision-tree-based
  classification of lifetime maximum intensity of tropical cyclones in the
  tropical western north pacific, Atmosphere 12~(7) (2021) 802.

\bibitem{zhao2019power}
W.~Zhao, L.~Shang, J.~Sun, Power quality disturbance classification based on
  time-frequency domain multi-feature and decision tree, Protection and Control
  of Modern Power Systems 4~(1) (2019) 1--6.

\bibitem{giniequa}
link, Decision tree available online:
  https://ekamperi.github.io/machine20learning/2021/04/13/gini-index-vs-entropy-decision-trees.html.

\bibitem{peng2018intrusion}
K.~Peng, V.~Leung, L.~Zheng, S.~Wang, C.~Huang, T.~Lin, Intrusion detection
  system based on decision tree over big data in fog environment, Wireless
  Communications and Mobile Computing 2018 (2018).

\bibitem{cervantes2020comprehensive}
J.~Cervantes, F.~Garcia-Lamont, L.~Rodr{\'\i}guez-Mazahua, A.~Lopez, A
  comprehensive survey on support vector machine classification: Applications,
  challenges and trends, Neurocomputing 408 (2020) 189--215.

\bibitem{vijayarajeswari2019classification}
R.~Vijayarajeswari, P.~Parthasarathy, S.~Vivekanandan, A.~A. Basha,
  Classification of mammogram for early detection of breast cancer using svm
  classifier and hough transform, Measurement 146 (2019) 800--805.

\bibitem{toledo2019support}
D.~C. Toledo-P{\'e}rez, J.~Rodr{\'\i}guez-Res{\'e}ndiz, R.~A. G{\'o}mez-Loenzo,
  J.~Jauregui-Correa, Support vector machine-based emg signal classification
  techniques: A review, Applied Sciences 9~(20) (2019) 4402.

\bibitem{chandra2021survey}
M.~A. Chandra, S.~Bedi, Survey on svm and their application in image
  classification, International Journal of Information Technology 13~(5) (2021)
  1--11.

\bibitem{pal2010feature}
M.~Pal, G.~M. Foody, Feature selection for classification of hyperspectral data
  by svm, IEEE Transactions on Geoscience and Remote Sensing 48~(5) (2010)
  2297--2307.

\bibitem{bazi2006toward}
Y.~Bazi, F.~Melgani, Toward an optimal svm classification system for
  hyperspectral remote sensing images, IEEE Transactions on geoscience and
  remote sensing 44~(11) (2006) 3374--3385.

\bibitem{hartmann2019comparing}
J.~Hartmann, J.~Huppertz, C.~Schamp, M.~Heitmann, Comparing automated text
  classification methods, International Journal of Research in Marketing 36~(1)
  (2019) 20--38.

\bibitem{deng2019feature}
X.~Deng, Y.~Li, J.~Weng, J.~Zhang, Feature selection for text classification: A
  review, Multimedia Tools and Applications 78~(3) (2019) 3797--3816.

\bibitem{churcher2021experimental}
A.~Churcher, R.~Ullah, J.~Ahmad, S.~Ur~Rehman, F.~Masood, M.~Gogate,
  F.~Alqahtani, B.~Nour, W.~J. Buchanan, An experimental analysis of attack
  classification using machine learning in iot networks, Sensors 21~(2) (2021)
  446.

\bibitem{mukherjee2012intrusion}
S.~Mukherjee, N.~Sharma, Intrusion detection using naive bayes classifier with
  feature reduction, Procedia Technology 4 (2012) 119--128.

\bibitem{komkov2021rf}
H.~Komkov, L.~Pocher, A.~Restelli, B.~Hunt, D.~Lathrop, Rf signal
  classification using boolean reservoir computing on an fpga, in: 2021
  International Joint Conference on Neural Networks (IJCNN), IEEE, 2021, pp.
  1--9.

\bibitem{peng2018design}
L.~Peng, A.~Hu, J.~Zhang, Y.~Jiang, J.~Yu, Y.~Yan, Design of a hybrid rf
  fingerprint extraction and device classification scheme, IEEE Internet of
  Things Journal 6~(1) (2018) 349--360.

\bibitem{sheykhmousa2020support}
M.~Sheykhmousa, M.~Mahdianpari, H.~Ghanbari, F.~Mohammadimanesh, P.~Ghamisi,
  S.~Homayouni, Support vector machine versus random forest for remote sensing
  image classification: A meta-analysis and systematic review, IEEE Journal of
  Selected Topics in Applied Earth Observations and Remote Sensing 13 (2020)
  6308--6325.

\bibitem{ezuma2019detection}
M.~Ezuma, F.~Erden, C.~K. Anjinappa, O.~Ozdemir, I.~Guvenc, Detection and
  classification of uavs using rf fingerprints in the presence of wi-fi and
  bluetooth interference, IEEE Open Journal of the Communications Society 1
  (2019) 60--76.

\bibitem{catic2018application}
A.~Catic, L.~Gurbeta, A.~Kurtovic-Kozaric, S.~Mehmedbasic, A.~Badnjevic,
  Application of neural networks for classification of patau, edwards, down,
  turner and klinefelter syndrome based on first trimester maternal serum
  screening data, ultrasonographic findings and patient demographics, BMC
  medical genomics 11~(1) (2018) 1--12.

\bibitem{arulmurugan2018early}
R.~Arulmurugan, H.~Anandakumar, Early detection of lung cancer using wavelet
  feature descriptor and feed forward back propagation neural networks
  classifier, in: Computational vision and bio inspired computing, Springer,
  2018, pp. 103--110.

\bibitem{yang2019feed}
J.~Yang, J.~Ma, Feed-forward neural network training using sparse
  representation, Expert Systems with Applications 116 (2019) 255--264.

\bibitem{statistics-FNN}
K.~Kamali,
  \href{\url{https://training.galaxyproject.org/training-material/topics/statistics/tutorials/FNN/tutorial.html}}{Deep
  learning (part 1) - feedforward neural networks (fnn) (galaxy training
  materials)}, [Online; accessed Wed Mar 16 2022] (06 2021).
\newline\urlprefix\url{\url{https://training.galaxyproject.org/training-material/topics/statistics/tutorials/FNN/tutorial.html}}

\bibitem{nagabushanam2020eeg}
P.~Nagabushanam, S.~Thomas~George, S.~Radha, Eeg signal classification using
  lstm and improved neural network algorithms, Soft Computing 24~(13) (2020)
  9981--10003.

\bibitem{jang2020bi}
B.~Jang, M.~Kim, G.~Harerimana, S.-u. Kang, J.~W. Kim, Bi-lstm model to
  increase accuracy in text classification: Combining word2vec cnn and
  attention mechanism, Applied Sciences 10~(17) (2020) 5841.

\bibitem{yildirim2019new}
O.~Yildirim, U.~B. Baloglu, R.-S. Tan, E.~J. Ciaccio, U.~R. Acharya, A new
  approach for arrhythmia classification using deep coded features and lstm
  networks, Computer methods and programs in biomedicine 176 (2019) 121--133.

\bibitem{saadatnejad2019lstm}
S.~Saadatnejad, M.~Oveisi, M.~Hashemi, Lstm-based ecg classification for
  continuous monitoring on personal wearable devices, IEEE journal of
  biomedical and health informatics 24~(2) (2019) 515--523.

\bibitem{lu2021review}
J.~Lu, L.~Tan, H.~Jiang, Review on convolutional neural network (cnn) applied
  to plant leaf disease classification, Agriculture 11~(8) (2021) 707.

\bibitem{dai2020hs}
G.~Dai, J.~Zhou, J.~Huang, N.~Wang, Hs-cnn: a cnn with hybrid convolution scale
  for eeg motor imagery classification, Journal of neural engineering 17~(1)
  (2020) 016025.

\bibitem{phan2018joint}
H.~Phan, F.~Andreotti, N.~Cooray, O.~Y. Ch{\'e}n, M.~De~Vos, Joint
  classification and prediction cnn framework for automatic sleep stage
  classification, IEEE Transactions on Biomedical Engineering 66~(5) (2018)
  1285--1296.

\bibitem{yu2020simplified}
C.~Yu, R.~Han, M.~Song, C.~Liu, C.-I. Chang, A simplified 2d-3d cnn
  architecture for hyperspectral image classification based on
  spatial--spectral fusion, IEEE Journal of Selected Topics in Applied Earth
  Observations and Remote Sensing 13 (2020) 2485--2501.

\bibitem{khourdifi2019heart}
Y.~Khourdifi, M.~Bahaj, Heart disease prediction and classification using
  machine learning algorithms optimized by particle swarm optimization and ant
  colony optimization, International Journal of Intelligent Engineering and
  Systems 12~(1) (2019) 242--252.

\bibitem{nguyen2008survey}
T.~T. Nguyen, G.~Armitage, A survey of techniques for internet traffic
  classification using machine learning, IEEE communications surveys \&
  tutorials 10~(4) (2008) 56--76.

\bibitem{NSL-KDDdataset}
dataset link, {NSL-KDD dataset}. available on:
  http://nsl.cs.unb.ca/kdd/nsl-kdd.html, march 2009.

\bibitem{tavallaee2009detailed}
M.~Tavallaee, E.~Bagheri, W.~Lu, A.~A. Ghorbani, A detailed analysis of the kdd
  cup 99 data set, in: 2009 IEEE symposium on computational intelligence for
  security and defense applications, IEEE, 2009, pp. 1--6.

\bibitem{ingre2015performance}
B.~Ingre, A.~Yadav, Performance analysis of nsl-kdd dataset using ann, in: 2015
  international conference on signal processing and communication engineering
  systems, IEEE, 2015, pp. 92--96.

\bibitem{van2008visualizing}
L.~Van~der Maaten, G.~Hinton, Visualizing data using t-sne., Journal of machine
  learning research 9~(11) (2008).

\bibitem{pang2019signature}
Y.~Pang, Z.~Chen, L.~Peng, K.~Ma, C.~Zhao, K.~Ji, A signature-based assistant
  random oversampling method for malware detection, in: 2019 18th IEEE
  International conference on trust, security and privacy in computing and
  communications/13th IEEE international conference on big data science and
  engineering (TrustCom/BigDataSE), IEEE, 2019, pp. 256--263.

\bibitem{sharma2017pros}
G.~Sharma, Pros and cons of different sampling techniques, International
  journal of applied research 3~(7) (2017) 749--752.

\bibitem{scikit-learn}
link, Scikit-learn available online:
  https://scikit-learn.org/dev/versions.html.

\bibitem{patro2014augmenting}
V.~M. Patro, M.~R. Patra, Augmenting weighted average with confusion matrix to
  enhance classification accuracy, Transactions on Machine Learning and
  Artificial Intelligence 2~(4) (2014) 77--91.

\bibitem{}
in: Proceedings of.

\bibitem{Ever-2019}
Y.~K. Ever, B.~Sekeroglu, K.~Dimililer, Classification analysis of intrusion
  detection on {NSL-KDD} using machine learning algorithms, in: Proceedings of
  International Conference on Mobile Web and Intelligent Information Systems,
  Lecture Notes in Computer Science, vol 11673, Springer, Cham, 2019.
\newblock \href {https://doi.org//10.1007/978-3-030-27192-3\_9}
  {\path{doi:/10.1007/978-3-030-27192-3\_9}}.

\bibitem{XianweiGao-2019}
X.~Gao, C.~Shan, C.~Hu, Z.~Niu, Z.~Liu, An adaptive ensemble machine learning
  model for intrusion detection, IEEE Access 7 (2019) 82512 -- 82521.
\newblock \href {https://doi.org/10.1109/ACCESS.2019.2923640}
  {\path{doi:10.1109/ACCESS.2019.2923640}}.

\bibitem{Shone-2018}
N.~Shone, T.~N. Ngoc, V.~D. Phai, Q.~Shi, A deep learning approach to network
  intrusion detection, IEEE Transactions on Emerging Topics in Computational
  Intelligence 2~(1) (2018) 41--50.
\newblock \href {https://doi.org/10.1109/TETCI.2017.2772792}
  {\path{doi:10.1109/TETCI.2017.2772792}}.

\bibitem{yin2017deep}
C.~Yin, Y.~Zhu, J.~Fei, X.~He, A deep learning approach for intrusion detection
  using recurrent neural networks, IEEE Access 5 (2017) 21954--21961.

\bibitem{javaid2016deep}
A.~Javaid, Q.~Niyaz, W.~Sun, M.~Alam, A deep learning approach for network
  intrusion detection system, in: Proceedings of the 9th EAI International
  Conference on Bio-inspired Information and Communications Technologies
  (formerly BIONETICS), 2016, pp. 21--26.

\bibitem{shorten2019survey}
C.~Shorten, T.~M. Khoshgoftaar, A survey on image data augmentation for deep
  learning, Journal of big data 6~(1) (2019) 1--48.

\bibitem{li2020intelligent}
X.~Li, W.~Zhang, Q.~Ding, J.-Q. Sun, Intelligent rotating machinery fault
  diagnosis based on deep learning using data augmentation, Journal of
  Intelligent Manufacturing 31~(2) (2020) 433--452.

\bibitem{zhou2020forecasting}
Y.~Zhou, F.~Dong, Y.~Liu, Z.~Li, J.~Du, L.~Zhang, Forecasting emerging
  technologies using data augmentation and deep learning, Scientometrics
  123~(1) (2020) 1--29.

\bibitem{moustafa2015UNSW}
N.~Moustafa, J.~Slay, {UNSW-NB15: A} comprehensive data set for network
  intrusion detection systems ({UNSW-NB}15 network data set), in: Proceedings
  of 2015 military communications and information systems conference (MilCIS),
  IEEE, 2015, pp. 1--6.

\bibitem{moustafa2016evaluation}
N.~Moustafa, J.~Slay, The evaluation of network anomaly detection systems:
  Statistical analysis of the {UNSW-NB}15 data set and the comparison with the
  {KDD}99 data set, Information Security Journal: A Global Perspective 25~(1-3)
  (2016) 18--31.

\bibitem{sharafaldin2018toward}
I.~Sharafaldin, A.~H. Lashkari, A.~A. Ghorbani, Toward generating a new
  intrusion detection dataset and intrusion traffic characterization., ICISSp 1
  (2018) 108--116.

\end{thebibliography}
\end{document}